\documentclass[preprint,12pt,nopreprintline]{elsarticle}

\usepackage{iftex}
\ifPDFTeX
  \usepackage[T1]{fontenc}
  \usepackage[utf8]{inputenc}
\else
  \usepackage{fontspec}
\fi

\usepackage{xspace}
\newcommand{\our}{PALATE}

\usepackage{amsmath}
\usepackage{amssymb}
\usepackage{mathtools}
\usepackage{amsthm}
\usepackage{dsfont}
\usepackage{xcolor}

\newtheorem{thm}{Theorem}
\newtheorem{df}{Definition}
\renewcommand{\P}{\mathbb{P}}
\newcommand{\E}{\mathbb{E}}

\usepackage{graphicx}
\usepackage{multirow}
\usepackage{makecell}
\usepackage{wrapfig}
\usepackage{fancyvrb}
\usepackage{url}
\usepackage{subcaption}
\usepackage{booktabs}
\usepackage{colortbl}
\usepackage{array}
\definecolor{negblue}{RGB}{13,105,172}
\definecolor{posred}{RGB}{211,47,47}

\usepackage[hidelinks]{hyperref}
\hypersetup{pdfauthor={Tadeusz Dziarmaga; Marcin Kądziołka; Laura Musioł; Patryk Marszałek; Artur Kasymov; Piotr Kościelniak; Marcin Mazur}}
\usepackage[capitalize,noabbrev]{cleveref}

\begin{document}
\frenchspacing

\begin{frontmatter}

\title{PALATE: Peculiar Application of the Law of Total Expectation to
       Enhance the Evaluation of Deep Generative Models}

\author{Tadeusz Dziarmaga\textsuperscript{*}}

\author{Marcin Kądziołka}

\author{Laura Musioł}

\author{Patryk Marszałek}

\author{Artur Kasymov}

\author{Piotr Kościelniak}

\author{Marcin Mazur\textsuperscript{*,\ensuremath{\dagger}}}

\nonumnote{\raggedright
\textsuperscript{*}These authors contributed equally to this work.\\
\textsuperscript{\ensuremath{\dagger}}Corresponding author: \texttt{marcin.mazur@uj.edu.pl}.}

\affiliation{organization={Faculty of Mathematics and Computer Science,
                            Jagiellonian University},
                city={Kraków},
                country={Poland}}

\begin{abstract}
\small


Deep generative models (DGMs) have caused a paradigm shift in the field of machine learning, yielding noteworthy advancements in domains such as image synthesis, natural language processing, and other related areas. However, a comprehensive evaluation of these models that effectively captures the interplay among fidelity, diversity, and novelty in generated samples remains a persistent challenge. A recently proposed approach, the Feature Likelihood Divergence (FLD), offers a theoretically grounded and practical tool in this regard but faces notable computational limitations. In this paper, we propose PALATE, a novel framework designed to enhance the evaluation of DGMs by addressing the efficiency constraints of FLD. Our approach is based on a peculiar application of the law of total expectation to random variables representing accessible data. When integrated with the MMD baseline metric and the recently introduced DINOv3 feature extractor, PALATE provides a comprehensive evaluation framework that matches or exceeds state-of-the-art methods while offering superior computational efficiency and scalability to large-scale datasets. Through a series of experiments, we demonstrate that the PALATE enhancement constitutes a practical and computationally efficient framework for holistic DGM assessment, particularly in detecting sample memorization and evaluating generalization capabilities. The source~code is provided as supplementary material.
\end{abstract}


\end{frontmatter}

\section{Introduction}
\label{sec:intro}

In recent years, deep generative models (DGMs) have garnered significant attention, drawing interest not only from the machine learning community but also from researchers in other scientific disciplines, practitioners, and the general public~\citep{ravuri2023understanding}. DGMs have matured to the point of enabling strong downstream results, including photorealistic imagery~\citep{rombach2022high}, speech that emulates authentic discourse~\citep{goel2022s}, and text reminiscent of human writing~\citep{openai2023gpt}.

\begin{figure}[t]
 \vskip2mm 
  \includegraphics[width=\linewidth]{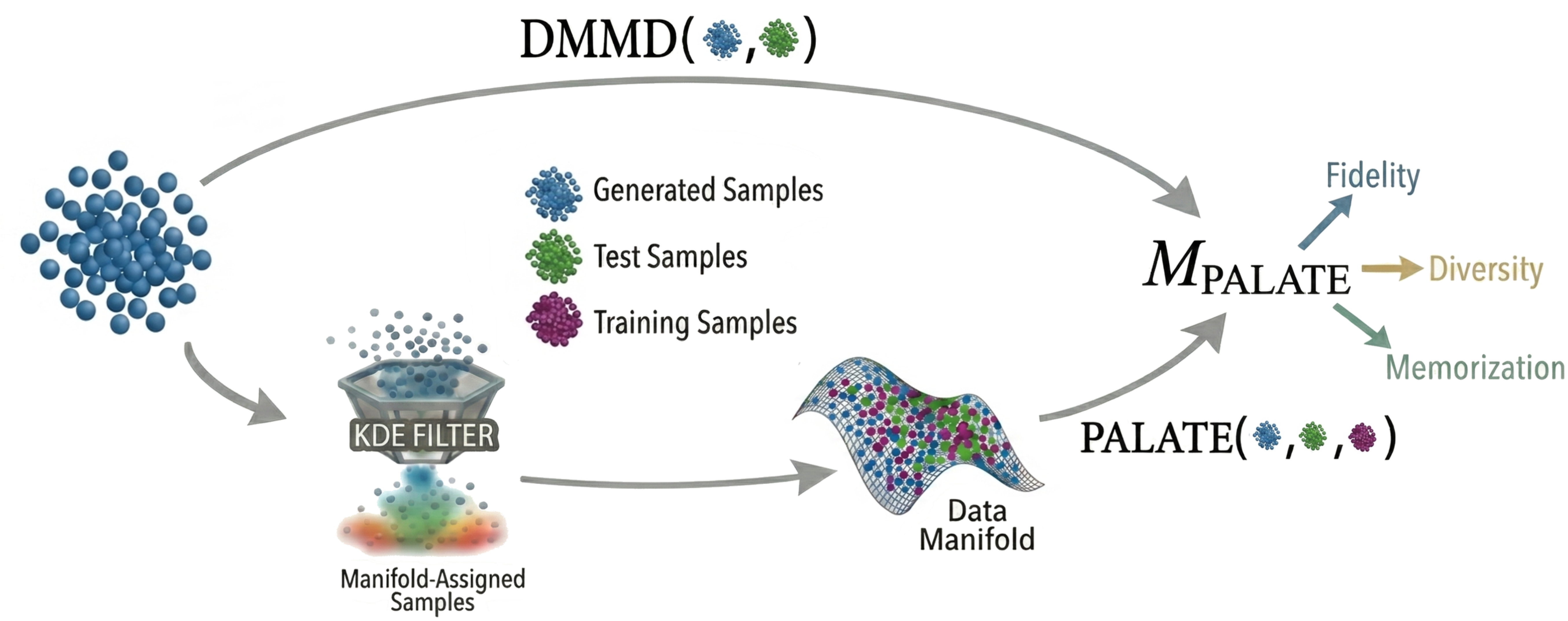}
  
  \caption{We propose PALATE, a memorization-aware enhancement to deep generative model assessment. Grounded in a peculiar application of the law of total expectation and leveraging DINOv3 representations, it yields $M_{\text{PALATE}}$, an MMD-based holistic evaluation metric that captures fidelity, diversity, and memorization in a single score.}
  \label{fig:teaser}
\end{figure}

Regardless of architecture, DGMs are trained to generate synthetic data that closely resemble real data. Since learning objectives are model-specific, fair post-training assessment is typically architecture agnostic and therefore sample-based, comparing generated samples to real data. The most prominent examples are Inception Score (IS)~\citep{salimans2016improved} and Fr\'echet Inception Distance (FID)~\citep{heusel2017gans}, popular due to usability, reasonable agreement with human perceptual judgments, and sensitivity to diversity. In particular, FID often captures fidelity by measuring a statistical distance between feature distributions fitted to real and generated samples. Nevertheless, despite its de facto role as a gold standard, FID generally lacks the capacity to detect overfitting or sample memorization~\citep{jiralerspong2024feature}.
More recent metrics such as Kernel Inception Distance (KID)~\citep{binkowski2018demystifying} or Precision and Recall (P\&R)~\citep{sajjadi2018assessing} address other weaknesses of FID, such as bias, normality assumptions, or the inability to separate fidelity from diversity, but they still do not resolve memorization sensitivity. Complementary memorization metrics, including the $C_T$ Score~\citep{meehan2020non} and Authenticity~\citep{alaa2022faithful}, target overfitting directly by quantifying training data reuse. While important, this separation of fidelity and diversity evaluation from memorization evaluation complicates model selection, especially in large-scale benchmarking where computational efficiency and a unified, actionable signal are critical.

A desirable evaluation metric should therefore be holistic, providing a single value that jointly validates generated samples in terms of fidelity, diversity, and novelty, where novelty is the counterpart of memorization. A recent and promising step is Feature Likelihood Divergence (FLD)~\citep{jiralerspong2024feature}, designed to assess generalization while preserving benefits of feature-based evaluation. Nevertheless, FLD can become computationally challenging on complex, large-scale datasets, because reliable estimation may require substantial numbers of samples (see our ImageNet results in \cref{sec:experiments}).
To address this issue, we propose a novel enhancement to the evaluation of DGMs, which we refer to as \our{}. Our solution is grounded in a \textbf{p}eculiar \textbf{a}pplication of the \textbf{la}w of \textbf{t}otal \textbf{e}xpectation (hence the name). The objective is to upgrade a baseline metric that already captures fidelity and diversity, making it sensitive to novelty and memorization while preserving the simplicity of sample-based evaluation (see \cref{fig:teaser} for an overview). Our experiments show that \our{}, when paired with an MMD-based approach, yields a single-value holistic evaluation metric that is aligned with state-of-the-art solutions while offering improved computational efficiency. To the best of our knowledge, our work is the first to incorporate the DINOv3 feature extractor~\cite{simeoni2025dinov3} into DGM evaluation.

In summary, our work contributes the following:
\begin{itemize}
\setlength{\emergencystretch}{1em}
\item We propose \our{}, a novel general enhancement for evaluating DGMs that upgrades a baseline metric to ensure sensitivity to training sample memorization while preserving fidelity and diversity assessment.
\item We provide a theoretical justification based on a peculiar application of the law of total expectation to random variables defined over a Euclidean data space.
\item We conduct extensive experiments on real-world datasets, introducing DINOv3 embeddings to DGM evaluation and showing that our approach yields a computationally efficient single-value holistic metric that matches, and sometimes exceeds, state-of-the-art solutions.
\end{itemize}

\section{Related Work}
\label{sec:related}

In this section, we review state-of-the-art evaluation metrics for DGMs that are relevant to our work.

\paragraph{Fidelity and Diversity Metrics}
Inception Score (IS)~\citep{salimans2016improved} is a sample-based metric that uses a pre-trained Inception-v3 network~\citep{szegedy2016rethinking} and primarily reflects diversity, with limited ability to assess fidelity. Metrics targeting both fidelity and diversity include Fr\'echet Inception Distance (FID)~\citep{heusel2017gans}, Kernel Inception Distance (KID)~\citep{binkowski2018demystifying}, and CLIP-MMD (CMMD)~\citep{jayasumana2024rethinking}. FID compares fitted feature distributions of real and generated samples, and it is sensitive both to biases induced by the embedding space and to the Gaussian approximation. In contrast, KID is MMD-based, avoids the Gaussian assumption, and is typically less biased. As part of a broader evaluation view, FD and KD~\citep{stein2024exposing} replace Inception features with DINOv2 embeddings, which can better capture global structure and salient objects, improving perceptual alignment. CMMD is also MMD-based, but uses CLIP embeddings and a Gaussian RBF kernel, differing from KID's typical Inception-v3 embeddings and kernel choices. Improved Precision and Recall~\citep{NEURIPS2019_0234c510} and Density and Coverage~\citep{naeem2020reliable} further decompose quality and coverage, providing more diagnostic feedback than a single scalar score. Beyond these, the Vendi Score~\citep{friedman2023vendi} offers a flexible, similarity-based notion of diversity that can expose mode collapse and duplication effects even when coverage appears high. Other recent proposals, such as FKEA~\citep{ospanov2024towards} and RKE Score~\citep{jalali2023information}, provide alternative, often kernel or information theoretic, perspectives on fidelity and diversity. However, these metrics are generally ineffective at detecting memorization and overfitting.

\paragraph{Memorization Metrics}
Concerns regarding DGMs memorizing training data have motivated metrics designed specifically to detect training data reuse. The $C_T$ Score~\citep{meehan2020non} quantifies the probability that a generated sample is indistinguishable from a real sample, while Authenticity~\citep{alaa2022faithful} applies a binary per-sample test and is commonly summarized as AuthPct, the percentage of generated samples classified as authentic. In this context, it is also useful to consider the Generalization Gap FLD \cite{jiralerspong2024feature}, obtained by subtracting FLD from its train-set counterpart, which serves as a memorization-oriented diagnostic. Novelty assessment has further benefited from metrics such as FINC~\citep{zhang2025unveiling}, while complementary metrics like the Rarity Score~\citep{han2022rarityscorenew} and KEN Score~\citep{zhang2024interpretableevaluationentropybasednovelty} quantify uniqueness and informational entropy of generated outputs relative to the training set. While these metrics provide important insights into memorization and overfitting, they do not fully capture, in a single score, the holistic quality of generated samples that also includes fidelity and diversity, motivating enhancements that consider all these aspects jointly.

\paragraph{Holistic Metrics}
Holistic evaluation aims to assess generalization by jointly considering fidelity, diversity, and novelty within a single scalar metric. Feature Likelihood Divergence (FLD)~\citep{jiralerspong2024feature}, which to our knowledge is the only efficient metric of this kind, integrates fidelity, diversity, and memorization via density estimation in a feature space, typically DINOv2, with bandwidth selection designed to penalize memorization. While effective, FLD can become computationally demanding on complex, large-scale datasets due to the large sample sizes required for reliable estimation.


\section{Preliminaries}\label{sec:preliminaries}

This section reviews the generative-model setting considered in this paper and recalls the law of total expectation, a key tool behind our method.

\paragraph{Generative Models}
We assume access to two disjoint collections of real data: a training set used for learning and a test set used exclusively for evaluation. Both can be viewed as independent draws from a random variable $X$ supported on a multidimensional Euclidean space $\mathcal{X}$. Following~\citep{bengio2013representation,loaiza2024deep}, we further assume that real data concentrate on a lower-dimensional manifold $\mathcal{M}_{\text{data}}\subset\mathcal{X}$, that is, $\P(X\in \mathcal{M}_{\text{data}})=1$.

Any deep generative model aims to produce synthetic samples. We model generated data as independent draws from a random variable $Y$ on the same space $\mathcal{X}$, with distribution $p_Y$ determined by the trained model. The real data distribution $p_X$ is unknown but fixed, whereas $p_Y$ evolves during training; learning aims to make $p_Y$ approximate $p_X$, typically by optimizing a model-specific loss on training samples. We emphasize that even for well-optimized models, a substantial fraction of generated samples may still fall outside $\mathcal{M}_{\text{data}}$. Nevertheless, we assume that $\P(Y\in \mathcal{M}_\text{data})>0$, which ensures that, by repeating the generation process sufficiently many times, we obtain samples lying on the data manifold.

To compare trained models fairly, we require an architecture-agnostic evaluation metric that can be computed from samples, either characterizing $p_Y$ alone (as IS) or measuring discrepancy between $p_X$ and $p_Y$ (as FID). Most modern metrics apply a feature extractor $\phi$, such as Inception-v3~\citep{szegedy2016rethinking}, DINOv2~\citep{oquab2023dinov2}, or CLIP~\citep{radford2021learning}, and evaluate the induced feature distributions $p_{\phi(X)}$ and $p_{\phi(Y)}$ in a perceptually meaningful representation space. The previous section reviewed the most relevant state-of-the-art evaluation metrics for our setting.

\paragraph{Law of Total Expectation}
The law of total expectation~\citep{kolmogorov1933foundations,feller1968introduction}, also known as the tower rule, expresses the expectation of a random variable as a weighted average of conditional expectations over a partition of the sample space. We state this form below.

\begin{thm}[\citep{kolmogorov1933foundations,feller1968introduction}]\label{thm:lote}
Let $Z$ be a random variable with finite expectation, and let $\{A_1,\ldots,A_n\}$ be a partition of a sample space $\Omega$, that is, $\bigcup_{i=1}^n A_i=\Omega$ and $A_i\cap A_j=\emptyset$ for $i\neq j$, with $\P(A_i)>0$ for all $i$. Then
\begin{equation}\label{eq:lote}
\E(Z)=\sum_{i=1}^n \E(Z\mid A_i)\P(A_i).
\end{equation}
\end{thm}

\section{\our{} Enhancement}

This section introduces \our{}, a novel enhancement for evaluating DGMs that builds on a peculiar application of the law of total expectation. Our goal is to endow a strong fidelity and diversity baseline with sensitivity to training-sample memorization, resulting in a single-score holistic evaluation metric $M_\text{PALATE}$.

\paragraph{PALATE}
 Let ${\rm x}_{\text{train}}=\{x^{\text{train}}_1,\ldots,x^{\text{train}}_m\}$ and ${\rm x}_{\text{test}}=\{x^{\text{test}}_1,\ldots,x^{\text{test}}_n\}$ be the training and test samples used for evaluation. We assume that these samples induce two disjoint, non-trivial regions on the data manifold\footnote{These train and test regions require no special structure (in particular, they need not be submanifolds or connected sets). They are induced solely by the samples selected for evaluation and are neither predetermined nor fixed subsets. 
 },
$\mathcal{M}_{\text{train}}$ and $\mathcal{M}_{\text{test}}$, such that
$\mathcal{M}_{\text{data}}=\mathcal{M}_{\text{train}}\cup \mathcal{M}_{\text{test}}$,
${\rm x}_{\text{train}}\subset \mathcal{M}_{\text{train}}$, and
${\rm x}_{\text{test}}\subset \mathcal{M}_{\text{test}}$.
We also assume that the ratio $m/n$ is fixed across all evaluated models and datasets.

Next, we consider a baseline evaluation metric $M_{\text{base}}$ that captures fidelity and diversity and that can be written as a conditional expectation $M_{\text{base}}=\E(Z\mid X\in \mathcal{M}_{\text{test}})$
for some random variable $Z=Z(X,Y)$ satisfying $\E(Z)=0$ if and only if $p_X=p_Y$ (see \cref{eq:mmd} for an example of such a construction). 
Applying Theorem~\ref{thm:lote} to the partition
$\Omega=\{\omega:X(\omega)\in \mathcal{M}_{\text{test}}\}\cup \{\omega:X(\omega)\in \mathcal{M}_{\text{train}}\}$
and conditioning on $Y\in \mathcal{M}_{\text{data}}$ yields
\begin{equation}\label{eq:lotetoz}
\begin{aligned}
\E(Z\mid Y\in \mathcal{M}_{\text{data}})
&= a\,\E(Z\mid X\in \mathcal{M}_{\text{test}},\,Y\in \mathcal{M}_{\text{data}})\\
&\quad + (1-a)\,\E(Z\mid X\in \mathcal{M}_{\text{train}},\,Y\in \mathcal{M}_{\text{data}}),
\end{aligned}
\end{equation}
where $a=\P(X\in \mathcal{M}_{\text{test}})=1-\P(X\in \mathcal{M}_{\text{train}})$ represents the probability mass assigned to the test region. Equation~\eqref{eq:lotetoz} is key: it decomposes the conditional expectation over the data manifold into a test-region term (aligned with standard fidelity--diversity evaluation) and a train-region term that captures the extent to which ``good'' samples can be explained by proximity to training data.

Motivated by this decomposition, we define
\begin{equation}\label{eq:palate}
\text{PALATE}
= \frac{a\,\E(Z\mid X\in \mathcal{M}_{\text{test}},\,Y\in \mathcal{M}_{\text{data}})}
{\E(Z\mid Y\in \mathcal{M}_{\text{data}})}.
\end{equation}
This ratio approaches $1$ for copycat models that resample training data and is minimized for models that generalize by producing novel samples consistent with the data distribution. Precisely, models with satisfactory generalization satisfy $\text{PALATE} \leq a$, in which case the contribution of generated samples to matching real data is largely explained by their proximity to the test set. In contrast, memorizing models satisfy $\text{PALATE} > a$, indicating that this contribution is less driven by test-data proximity than expected from the test-region mass alone.
To formalize this argument, inspired by~\citep{meehan2020non}, we introduce the following definition.

\begin{table}[!p]
\centering
    \caption{Evaluation of various DGMs on CIFAR-10 and ImageNet. Metrics implementations and experimental setups follow \citep{jiralerspong2024feature} (FLD and Gen. Gap FLD), \citep{jayasumana2024rethinking} (CMMD), and \citep{stein2024exposing} (all others). Calculations use the data examples from \citep{stein2024exposing}; mean per-class Vendi Score is omitted for some models due to missing class labels in the generated samples.}
    \label{tab:models}
{\scriptsize
\setlength{\tabcolsep}{2pt}
\scalebox{0.85}{%
\begin{tabular*}{1.2\textwidth}{@{\extracolsep{\fill}}ccccccccc@{}}
\toprule
      & & \multicolumn{2}{c}{Holistic metrics} & \multicolumn{3}{c}{Fidelity/diversity metrics} & \multicolumn{2}{c}{Fidelity metrics}\\
\cmidrule(rr){3-4} \cmidrule(rr){5-7} \cmidrule(rr){8-9}
 & Model & $M_\text{\our{}}$ (our)$\downarrow$ & FLD$\downarrow$ & FD$\downarrow$ & KD$\downarrow$ & CMMD$\downarrow$ & Precision$\uparrow$ & Density$\uparrow$ \\
\midrule
\multirow{13}{*}{\rotatebox{90}{CIFAR-10}} & LSGM-ODE & 0.2618 & 4.25 & 148.27 & 0.18 & 0.025 & 0.74 & 0.76 \\
 & PFGM++ & 0.2614 & 4.51 & 141.65 & 0.17 & 0.022 & 0.75 & 0.80 \\
 & iDDPM-DDIM & 0.2666 & 5.61 & 212.30 & 0.26 & 0.072 & 0.74 & 0.78 \\
 & StyleGAN-XL & 0.2652 & 5.52 & 204.60 & 0.24 & 0.018 & 0.72 & 0.69 \\
 & StyleGAN2-ada & 0.2721 & 6.91 & 305.92 & 0.45 & 0.037 & 0.71 & 0.70 \\
 & BigGAN-Deep & 0.2718 & 9.52 & 326.66 & 0.39 & 0.037 & 0.69 & 0.66 \\
 & ReACGAN & 0.2790 & 8.78 & 353.25 & 0.45 & 0.050 & 0.68 & 0.66 \\
 & MHGAN & 0.2802 & 8.73 & 358.07 & 0.49 & 0.049 & 0.70 & 0.69 \\
 & LOGAN & 0.3398 & 19.80 & 881.73 & 1.51 & 0.259 & 0.71 & 0.98 \\
 & WGAN-GP & 0.3624 & 23.66 & 1088.56 & 2.04 & 0.407 & 0.82 & 1.60 \\
 & ACGAN-Mod & 0.3770 & 25.86 & 1208.26 & 2.41 & 0.468 & 0.84 & 2.06 \\
 & NVAE & 0.3709 & 22.87 & 921.34 & 1.57 & 0.441 & 0.76 & 1.27 \\
 & RESFLOW & 0.3932 & 29.27 & 1301.38 & 2.64 & 0.617 & 0.87 & 1.91 \\
\midrule
\multirow{11}{*}{\rotatebox{90}{\makecell{ImageNet 256$\times$256}}} & DiT-XL-2-guided & 0.2604 & 1.43 & 99.16 & 0.03 & 0.212 & 0.99 & 1.31 \\
 & LDM & 0.2864 & 3.37 & 112.40 & 0.09 & 0.339 & 0.92 & 1.00 \\
 & DiT-XL-2 & 0.2882 & 2.56 & 79.36 & 0.06 & 0.306 & 0.93 & 1.01 \\
 & ADMG-ADMU & 0.2658 & 5.40 & 111.24 & 0.08 & 0.210 & 0.88 & 0.86 \\
 & ADM & 0.2796 & 8.93 & 203.45 & 0.19 & 0.315 & 0.80 & 0.67 \\
 & ADMG & 0.2664 & 5.83 & 123.27 & 0.09 & 0.208 & 0.87 & 0.80 \\
 & RQ-Transformer & 0.3539 & 11.86 & 304.05 & 0.34 & 0.547 & 0.76 & 0.69 \\
 & BigGAN & 0.3348 & 12.34 & 401.22 & 0.52 & 1.471 & 0.81 & 0.95 \\
 & Mask-GIT & 0.3142 & 7.24 & 214.45 & 0.24 & 0.866 & 0.84 & 0.77 \\
 & GigaGAN & 0.2894 & 8.45 & 214.88 & 0.24 & 0.348 & 0.81 & 0.73 \\
 & StyleGAN-XL & 0.3168 & 8.78 & 150.27 & 0.22 & 0.440 & 0.82 & 0.76 \\
\bottomrule
\end{tabular*}
}
\\[3mm]
\scalebox{0.85}{%
\begin{tabular*}{1.2\textwidth}{@{\extracolsep{\fill}}cccccccccc@{}}
\toprule
      & & \multicolumn{3}{c}{Diversity metrics} & \multicolumn{5}{c}{Memorization metrics}\\
\cmidrule(rr){3-5} \cmidrule(rr){6-10}
 & Model & Vendi$\downarrow$ & Recall$\uparrow$ & Coverage$\uparrow$ & \makecell{\our{} (our)$\downarrow$} & \makecell{G. G. FLD$\uparrow$} & $C_T$$\uparrow$ & \makecell{$C_T$ mod.$\uparrow$} & AuthPct$\uparrow$ \\
\midrule
\multirow{13}{*}{\rotatebox{90}{CIFAR-10}} & LSGM-ODE & --- & 0.74 & 0.71 & 0.5017 & -0.35 & 45.60 & -1.56 & 91.56 \\
 & PFGM++ & 86.18 & 0.71 & 0.72 & 0.5017 & -0.27 & 43.23 & -0.73 & 91.96 \\
 & iDDPM-DDIM & --- & 0.66 & 0.64 & 0.5018 & -0.21 & 48.64 & -0.89 & 92.13 \\
 & StyleGAN-XL & 87.95 & 0.55 & 0.60 & 0.5014 & -0.16 & 51.93 & -1.12 & 93.03 \\
 & StyleGAN2-ada & 97.50 & 0.51 & 0.52 & 0.5010 & -0.20 & 57.98 & -0.63 & 93.22 \\
 & BigGAN-Deep & 88.06 & 0.44 & 0.44 & 0.5013 & 0.01 & 62.62 & -0.21 & 94.04 \\
 & ReACGAN & 88.34 & 0.46 & 0.42 & 0.5014 & 0.04 & 62.87 & -0.46 & 93.49 \\
 & MHGAN & 87.79 & 0.47 & 0.41 & 0.5012 & 0.03 & 63.28 & -0.64 & 93.56 \\
 & LOGAN & 157.02 & 0.25 & 0.12 & 0.5000 & 0.08 & 69.40 & 0.10 & 91.51 \\
 & WGAN-GP & 112.02 & 0.04 & 0.08 & 0.5000 & 0.11 & 61.46 & -0.12 & 87.45 \\
 & ACGAN-Mod & 44.30 & 0.01 & 0.07 & 0.5000 & 0.09 & 59.90 & 0.56 & 82.38 \\
 & NVAE & --- & 0.27 & 0.11 & 0.5000 & 0.19 & 65.01 & 0.12 & 86.91 \\
 & RESFLOW & --- & 0.01 & 0.05 & 0.5000 & 0.16 & 49.17 & 0.02 & 76.72 \\
\midrule
\multirow{11}{*}{\rotatebox{90}{\makecell{ImageNet 256$\times$256}}} & DiT-XL-2-guided & 5.61 & 0.60 & 0.95 & 0.5014 & -1.42 & -43.29 & -5.16 & 61.46 \\
 & LDM & 14.17 & 0.76 & 0.89 & 0.5004 & -0.81 & 51.33 & -1.74 & 89.56 \\
 & DiT-XL-2 & 14.17 & 0.84 & 0.94 & 0.5005 & -1.02 & 34.32 & -3.25 & 85.46 \\
 & ADMG-ADMU & 15.09 & 0.84 & 0.86 & 0.5001 & -0.80 & 59.56 & -2.87 & 90.91 \\
 & ADM & 21.62 & 0.85 & 0.72 & 0.5001 & -0.41 & 104.43 & -2.03 & 93.34 \\
 & ADMG & 16.32 & 0.84 & 0.85 & 0.5004 & -0.63 & 70.19 & -1.73 & 92.35 \\
 & RQ-Transformer & 27.25 & 0.76 & 0.58 & 0.5001 & -0.37 & 124.25 & -1.61 & 94.30 \\
 & BigGAN & 12.78 & 0.39 & 0.54 & 0.4999 & -0.30 & 121.94 & -1.56 & 92.11 \\
 & Mask-GIT & 17.80 & 0.76 & 0.75 & 0.5002 & -0.69 & 97.40 & -1.58 & 92.60 \\
 & GigaGAN & 23.42 & 0.77 & 0.68 & 0.5002 & -0.49 & 106.24 & -2.73 & 93.37 \\
 & StyleGAN-XL & 16.75 & 0.72 & 0.68 & 0.5000 & -0.38 & 103.29 & -1.52 & 95.02 \\
\bottomrule
\end{tabular*}
}
}
\end{table} 



\begin{df}\label{def:datacopyingour}
A generative model is said to exhibit a tendency to memorization if 
\begin{equation}\label{eq:datacopyingour}
\!\!\E(Z \mid X\in \mathcal{M}_\textnormal{train}, Y\in \mathcal{M}_\text{data})
<
\E(Z \mid X\in \mathcal{M}_\textnormal{test}, Y\in \mathcal{M}_\text{data}).
\end{equation}
\end{df}

For models satisfying Definition~\ref{def:datacopyingour}, generated samples contribute more to matching the real data distribution through proximity to training data than to test data. Consequently, these are precisely the models for which the \our{} score exceeds $a = \mathbb{P}(X\in\mathcal{M}_\text{test})$, where we set $a=1/2$ in our implementation. Specifically,
\begin{equation}\label{eq:equiv}
\begin{aligned}
\E(Z \mid X\in \mathcal{M}_\text{train},  Y\in \mathcal{M}_\text{data}&)  <
\E(Z \mid X\in \mathcal{M}_\text{test}, Y\in \mathcal{M}_\text{data})\\
& \Longleftrightarrow\;
\text{PALATE}>a.
\end{aligned}
\end{equation}

\paragraph{\our{} Enhancement}
Although $\text{PALATE}$ explicitly measures memorization within $\mathcal{M}_{\text{data}}$, it may be ambiguous when comparing well-optimized and poorly optimized models, since the expectations in \Cref{eq:palate} are conditioned on $Y\in \mathcal{M}_{\text{data}}$. As a result, samples generated outside the data manifold do not affect $\text{PALATE}$. To retain sensitivity to such failures while still rewarding novelty, we define the final enhanced metric as a weighted combination of the baseline score (scaled to $[0,1]$) and $\text{PALATE}$:
\begin{equation}\label{eq:ourmetric}
M_{\text{PALATE}}
= \alpha\,\text{SCALE}(M_{\text{base}}) + (1-\alpha)\,\text{PALATE},
\end{equation}
where $\alpha\in[0,1)$. This construction retains the fidelity--diversity sensitivity of $M_{\text{base}}$ while adding memorization awareness through $\text{PALATE}$. Although not stated explicitly, this formulation involves hyperparameters and estimation procedures, which we discuss below.

\paragraph{Baseline Metric and Scaling Method}
We adopt as our baseline metric the Maximum Mean Discrepancy (MMD) \cite{gretton2006kernel} with the characteristic Gaussian RBF kernel
$k(x,y)=\exp(-\|x-y\|^2/(2\sigma^2))$.
This choice yields a distribution-free statistical distance in feature space:
\begin{equation}\label{eq:mmd}
\text{MMD}^2_k(p_{\phi(X)},p_{\phi(Y)})
= \E(k(X_1,X_2) + k(Y_1,Y_2) - 2k(X_1,Y_1)),
\end{equation}
where $X_1,X_2 \sim p_{\phi(X)}$ and $Y_1,Y_2 \sim p_{\phi(Y)}$ are independent. This formulation admits the following simple scaling method:
\begin{equation}\label{eq:scale}
\text{SCALE}(\text{MMD}^2_k)
= \frac{\text{MMD}^2_k(p_{\phi(X)},p_{\phi(Y)})}
{\E(k(X_1,X_2) + k(Y_1,Y_2))}
\in [0,1].
\end{equation}
The bandwidth $\sigma$ is tuned to balance fidelity, diversity, and memorization sensitivity, following the median heuristic of \cite{gretton2012kernel} (see \ref{app:impdet}).

We use DINOv3 as the feature extractor $\phi$ and denote the resulting baseline as DINOv3-MMD (DMMD).
We avoid Inception-v3 (used in KID) due to the bias induced by its supervised ImageNet training, and CLIP (used in CMMD) due to its limited fine-grained separability of object attributes \citep{bianchi2024clip}. Sensitivity to subtle features such as color and shape is essential for detecting overfitting and memorization. Prior work has shown that DINOv2 features enable more comprehensive DGM evaluation and correlate better with human judgment than Inception-v3 \citep{stein2024exposing}. DINOv3 further improves upon DINOv2 by providing denser and more stable patch-level features through Gram anchoring and large-scale training, yielding improved perceptual alignment and dense-prediction transfer \citep{simeoni2025dinov3}.

\paragraph{Data Manifold and Train--Test Split}
Since the data distribution $p_X$ is unknown, we approximate the data manifold using available real data samples by fitting a kernel density estimator (KDE) $\hat{f}(u \mid \phi({\rm x}_{\text{train}}\cup {\rm x}_{\text{test}}))$ in feature space.
We use a Gaussian kernel $K(u)=\mathcal{N}(u;0,\Sigma)$ with a bandwidth matrix
$\Sigma=\mathrm{diag}(\sigma_1^2,\ldots,\sigma_d^2)$, selected via Scott’s rule\footnote{Specifically, $\sigma_i = (m+n)^{-1/(d+4)} \hat{\sigma}_i$, where $d$ is the feature dimension and $\hat{\sigma}_i$ is the empirical standard deviation of the $i$th coordinate.}.
Under this representation, generated samples ${\rm y}=\{y_1,\ldots,y_k\}$ are assigned to the data manifold if they attain sufficiently high KDE log-likelihood:
\begin{equation}
y_i \in \mathcal{M}_{\text{data}}
\;\Leftrightarrow\;
\log \hat{f}(y_i \mid \phi({\rm x}_{\text{train}}\cup {\rm x}_{\text{test}})) > \tau,
\end{equation}
where the threshold $\tau$ is selected via grid search (see \ref{app:impdet}).

\begin{figure}[t]
\centering
\makebox[\textwidth][c]{\includegraphics[width=1.2\textwidth]{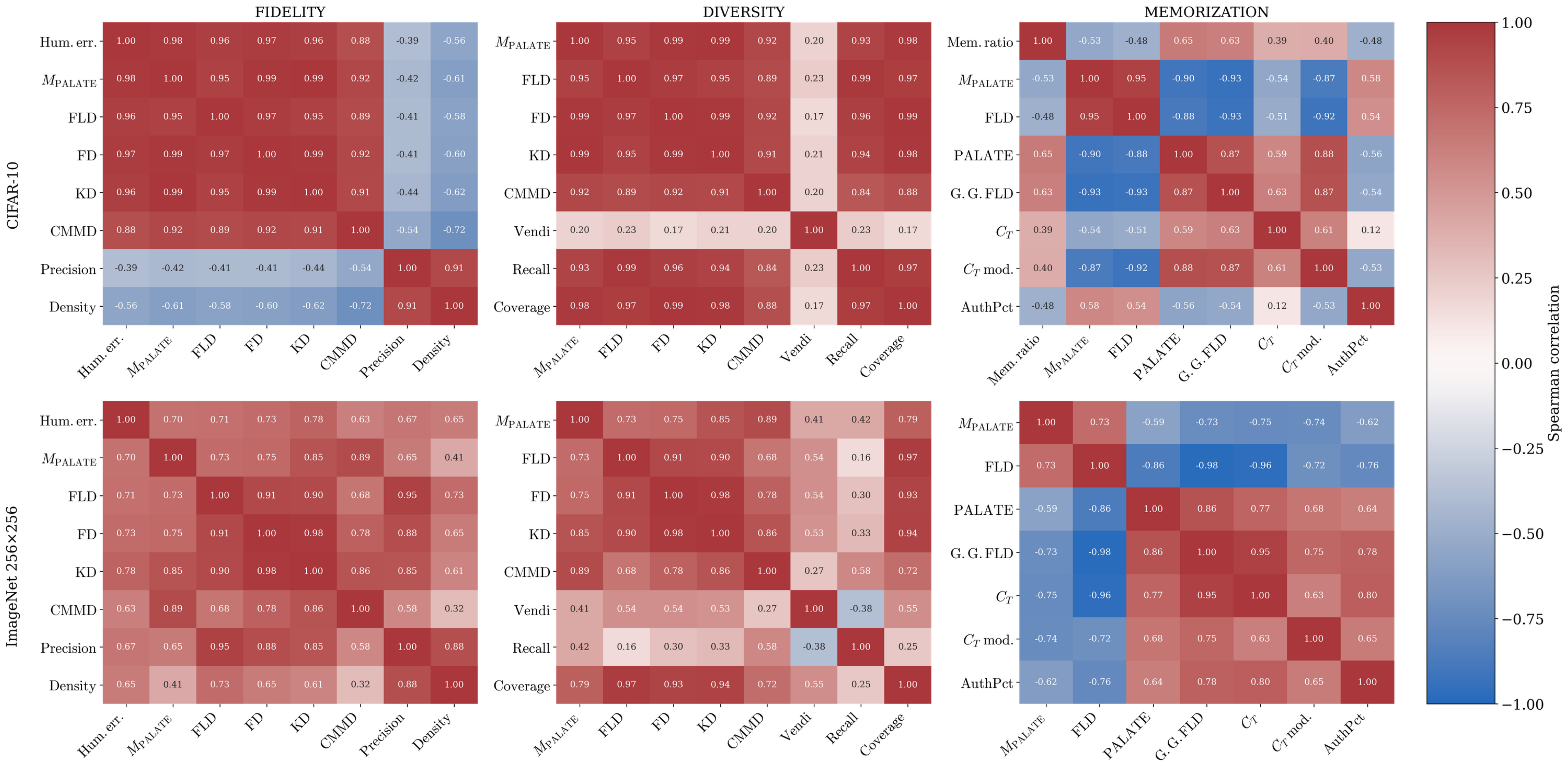}}
\caption{Spearman correlation heatmaps among evaluation metrics on CIFAR-10 (top row) and ImageNet (bottom row). For proper comparison, metric scales are aligned so that higher values consistently indicate better performance.}
\label{fig:metric-correlations}
\end{figure}

\begin{figure}[t]
\centering
\includegraphics[width=0.96\linewidth]{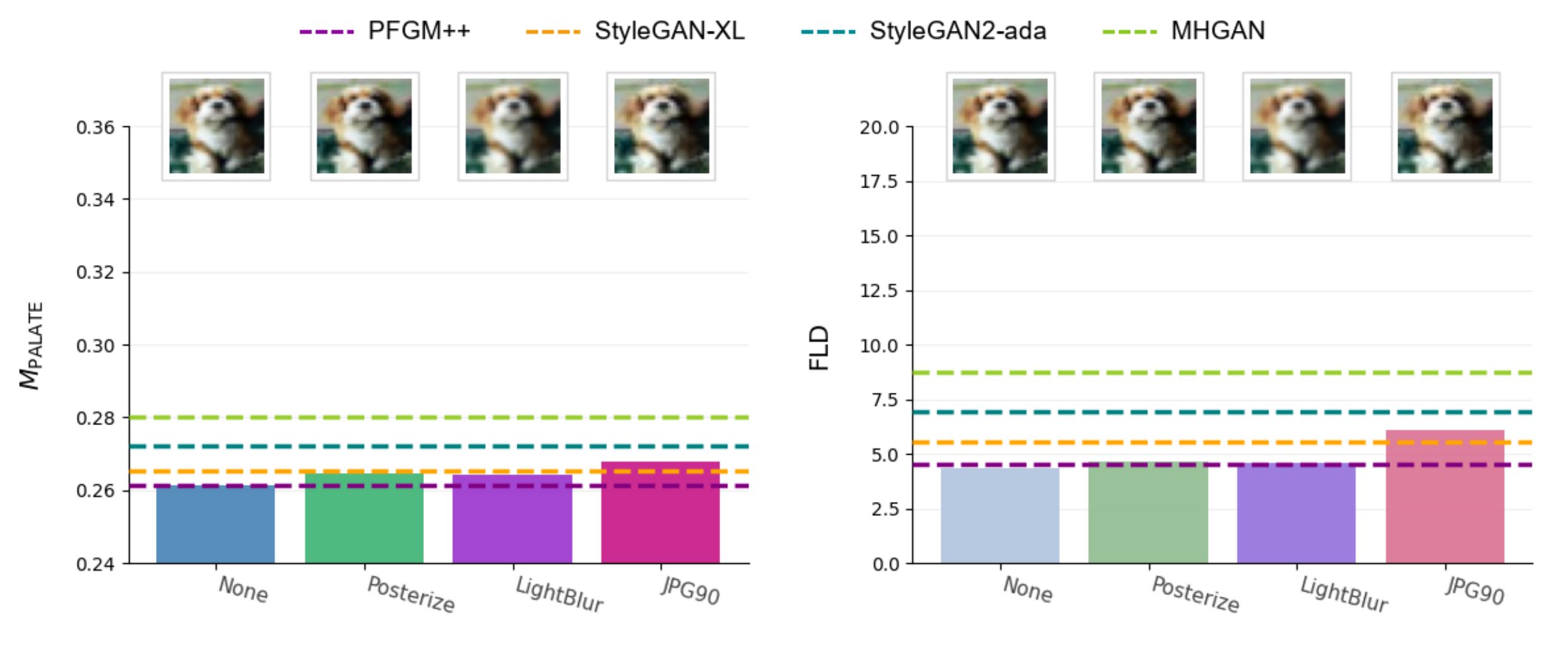}\\[-4mm]
\includegraphics[width=0.96\linewidth]{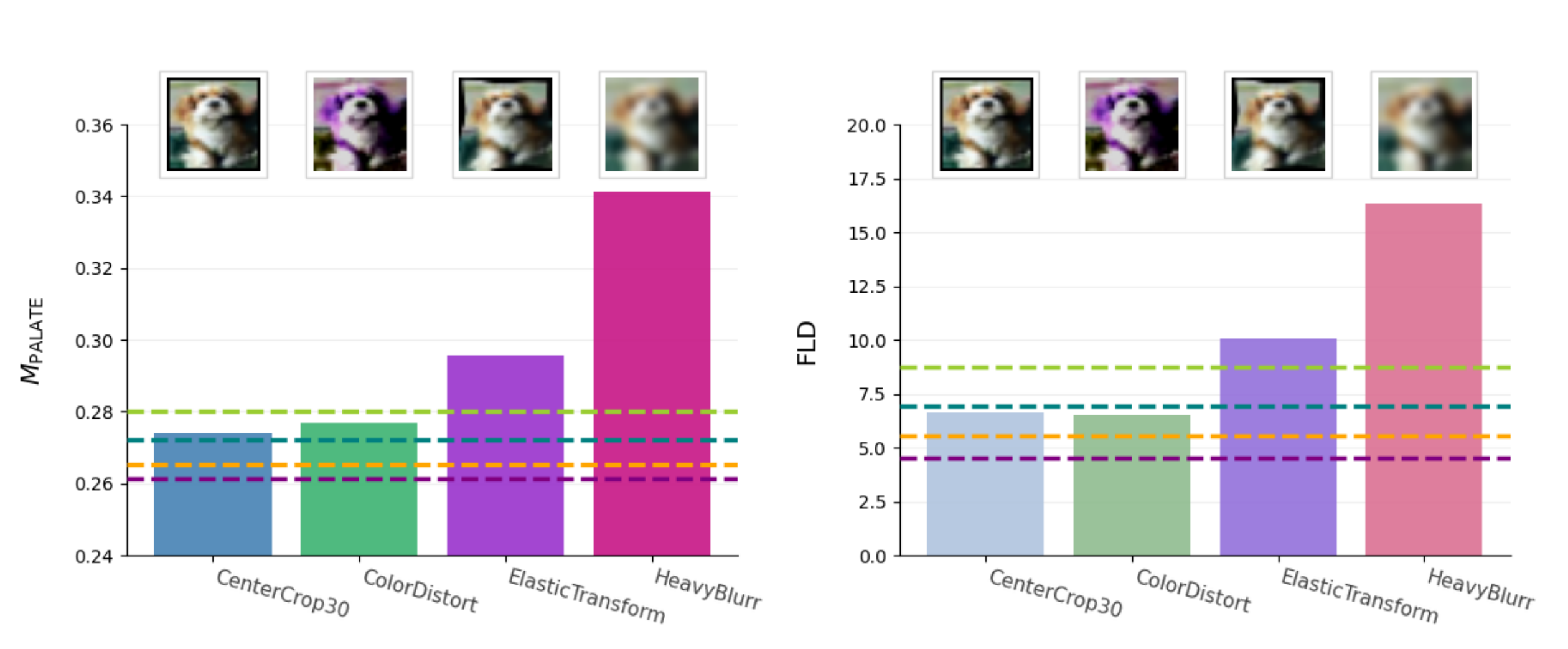}
\caption{ 
Comparison of $M_{\text{\our{}}}$ (our) and FLD under transformations applied to samples generated by PFGM++ on CIFAR-10. Results for other models are shown for reference.
{\em Top}: Nearly imperceptible transformations. {\em Bottom}: Large transformations. 
}
\label{fig:fidelity}
\end{figure}

We define $a=\mathbb{P}(X \in \mathcal{M}_{\text{test}})$ as the fraction of test samples among all real samples used for evaluation, i.e., $a = n/(m+n)$. Under our assumptions, this quantity remains fixed across models and datasets. This definition follows from an application of the law of total expectation to a one-dimensional embedding of $X$ (see \ref{sec:proofs}).

The remaining factors affecting computation are the sample sizes $m$, $n$, and $k$ for training, test, and generated data, respectively. We recommend using equal sizes ($m = n = k$), which is standard in DGM evaluation and fixes $a = 1/2$. Unlike most existing metrics, the PALATE enhancement explicitly incorporates training samples, aligning it with FLD. However, in contrast to our approach, FLD requires a sufficiently large training set; otherwise, the estimation of a dataset-dependent constant may result in negative values \citep{jiralerspong2024feature}. Moreover, FLD exhibits memory inefficiencies when processing a large number generated samples, further impacting its computational efficiency (see \cref{sec:experiments}).


\paragraph{Weighting Constant}
The weighting constant $\alpha$ controls the emphasis placed on memorization, thereby trading off sensitivity to fidelity and diversity in the generated samples. While $\alpha$ may be chosen arbitrarily depending on the evaluation goal, we focus on two representative settings: $\alpha = 0$ and $\alpha = 1/2$. The former yields a metric maximally sensitive to memorization, whereas the latter balances memorization with fidelity and diversity. Accordingly, our analysis centers on these two values of $\alpha$.

\begin{figure}[t]
    \centering
    \begin{subfigure}[t]{\columnwidth}
        \centering
        \includegraphics[width=\columnwidth]{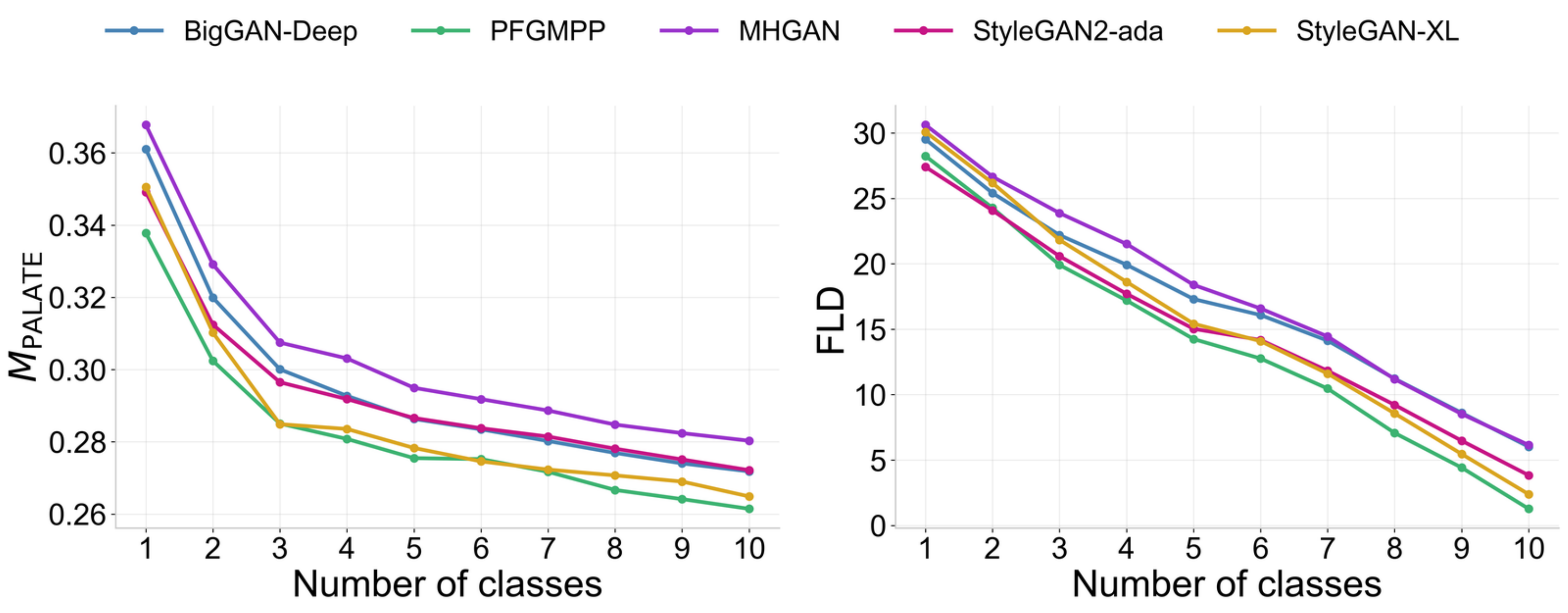}
        \label{fig:sample_div_classes}
    \end{subfigure}\\
    \begin{subfigure}[t]{\columnwidth}
        \centering
        \includegraphics[width=\columnwidth]{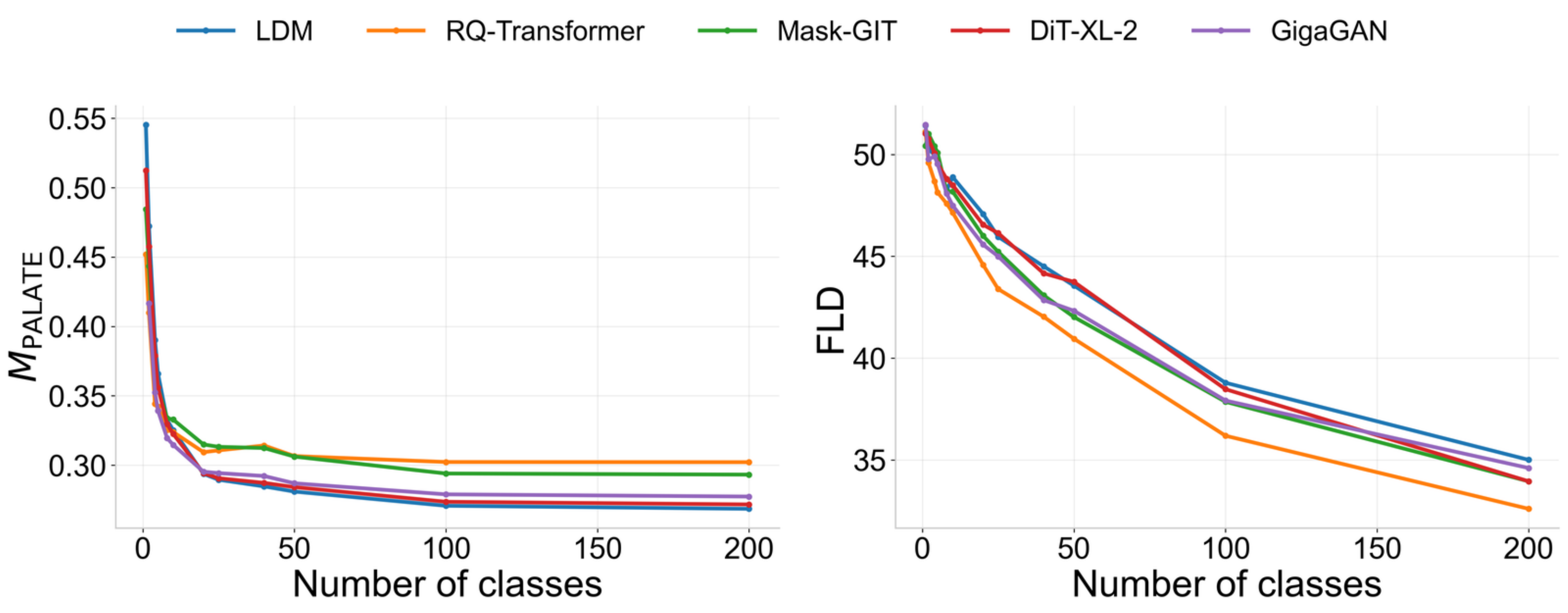}
        \label{fig:sample_div_unique}
    \end{subfigure}

    \caption{Capability of $M_{\text{\our{}}}$ (our) and FLD to capture increasing sample diversity as the number of classes grows, while keeping the total number of generated samples fixed at $10{,}000$.
{\em Top:} CIFAR-10.
{\em Bottom}: ImageNet.}
    \label{fig:sample_diverity}
\end{figure}

\begin{figure}[t] 
    \centering
    \includegraphics[width=1\columnwidth]{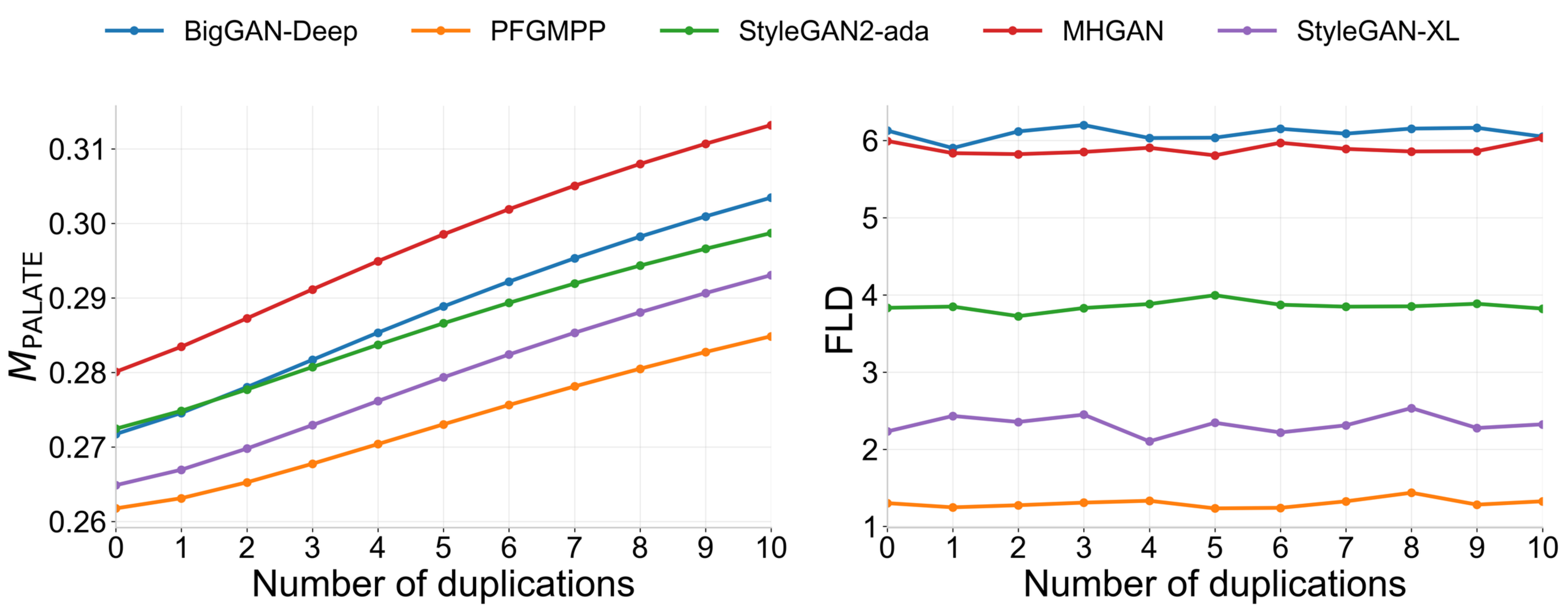}
    \caption{Capability of $M_{\text{\our{}}}$ (our) and FLD to detect diversity degradation under increasing class imbalance on CIFAR-10. We progressively duplicate $1{,}000$ fixed samples from a single (first) class while leaving all other classes unchanged.}
    \label{fig:class_imbalance}
\end{figure}

\paragraph{Metrics Computation}
As discussed above, we consider two evaluation metrics: $\text{PALATE}$, a memorization-focused metric, and $M_{\text{\our{}}}$, a single-score holistic metric. These are obtained from \cref{eq:ourmetric,eq:palate} by setting $\alpha = 0$ and $\alpha = 1/2$, respectively. We use $\text{DMMD}$ as the baseline metric, together with the $\text{SCALE}$ function defined in \cref{eq:scale}. Metric values are computed from real and generated samples ${\rm x}^{\text{train}}$, ${\rm x}^{\text{test}}$, and ${\rm y}$ by replacing all kernel expectations with their corresponding V-statistics, a standard approach in MMD estimation \citep{gretton2012kernel}.
 Explicit formulas for $\text{PALATE}$ and $M_{\text{\our{}}}$, along with implementation details, are provided in \ref{app:impdet}.

\section{Experiments}\label{sec:experiments}

In this section, we present the results of our experimental study on a range of deep generative models trained on two real-world datasets, CIFAR-10 and the large-scale ImageNet dataset\footnote{Both datasets provide an explicit separation between training and test examples.}. We also study a text-to-image overfitting scenario using a pre-trained Stable Diffusion v1.5 model fine-tuned on MS-COCO with LoRA~\cite{rombach2022high,hu2022lora}. Our experiments rely on data examples and implementations from \citep{stein2024exposing,jiralerspong2024feature,jayasumana2024rethinking,bai2021training}. Unless stated otherwise, we use $10{,}000$ training, test, and generated samples. While several baselines are considered, we primarily compare against FLD, which, to our knowledge, is the only state-of-the-art metric that efficiently captures fidelity, diversity, and novelty in a single score. Additional experimental results are provided in \ref{sec:add_exp}. The source~code is provided as supplementary material.

\paragraph{Evaluation of DGMs in Real-World Settings}
We evaluate $M_\text{\our{}}$ and \linebreak \our{} on CIFAR-10 and ImageNet using a broad set of DGMs and commonly used evaluation metrics. Raw metric values are reported in \cref{tab:models}, and \cref{fig:metric-correlations} shows Spearman’s $\rho$ rank-order correlation heatmaps\footnote{We use Spearman’s rank correlation to quantify agreement between evaluation metrics, as it robustly captures monotonic but potentially nonlinear relationships and is invariant to scale differences, unlike Pearson correlation, while being more stable and interpretable than Kendall’s $\tau$ for moderate sample sizes \cite{sajjadi2018assessing,li2024great,conover1999practical}.}. Following \cite{stein2024exposing}, we include Human Error Rate and Memorization Ratio as reference measures for fidelity and memorization, respectively (see \ref{app:impdet}).

\begin{figure}[t]
\centering

\begin{subfigure}[c]{0.92\textwidth}
    \centering
    \includegraphics[width=\linewidth,trim=0 6 0 0,clip]{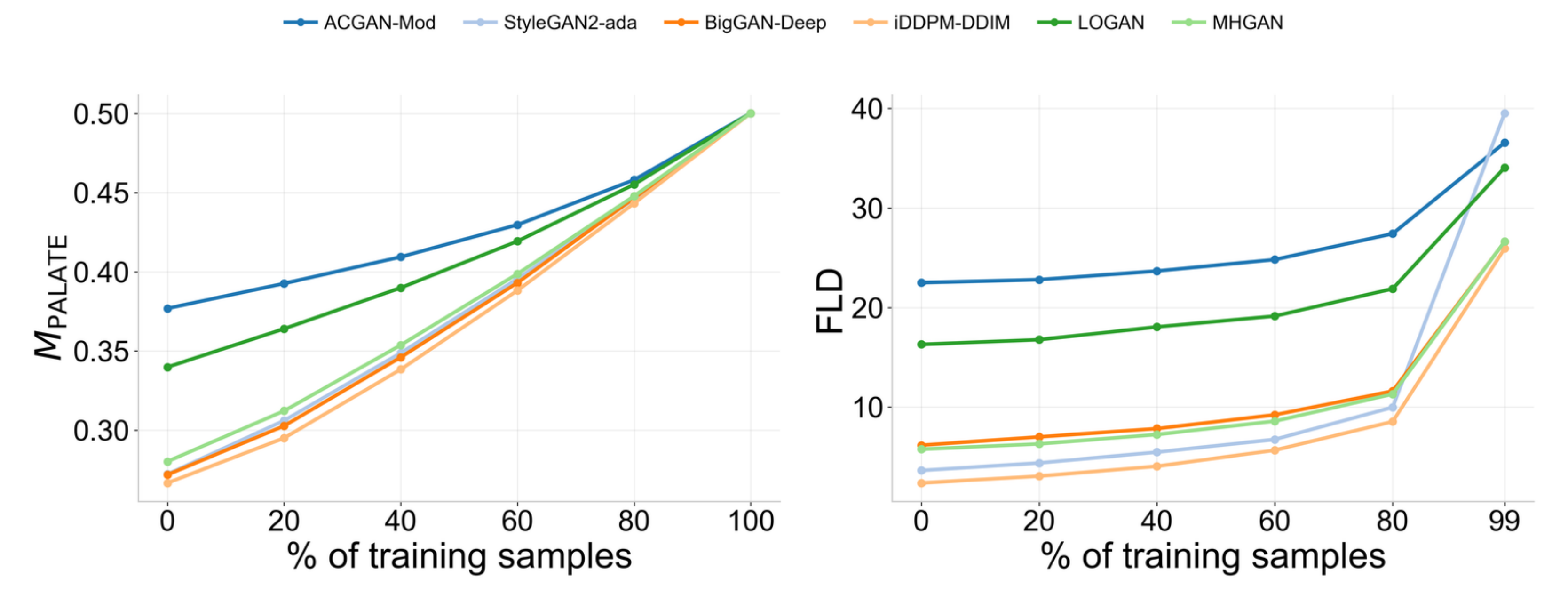}
\end{subfigure}\\[1.5mm]
\begin{subfigure}[c]{0.92\textwidth}
    \centering
    \includegraphics[width=\linewidth,trim=0 0 0 0,clip]{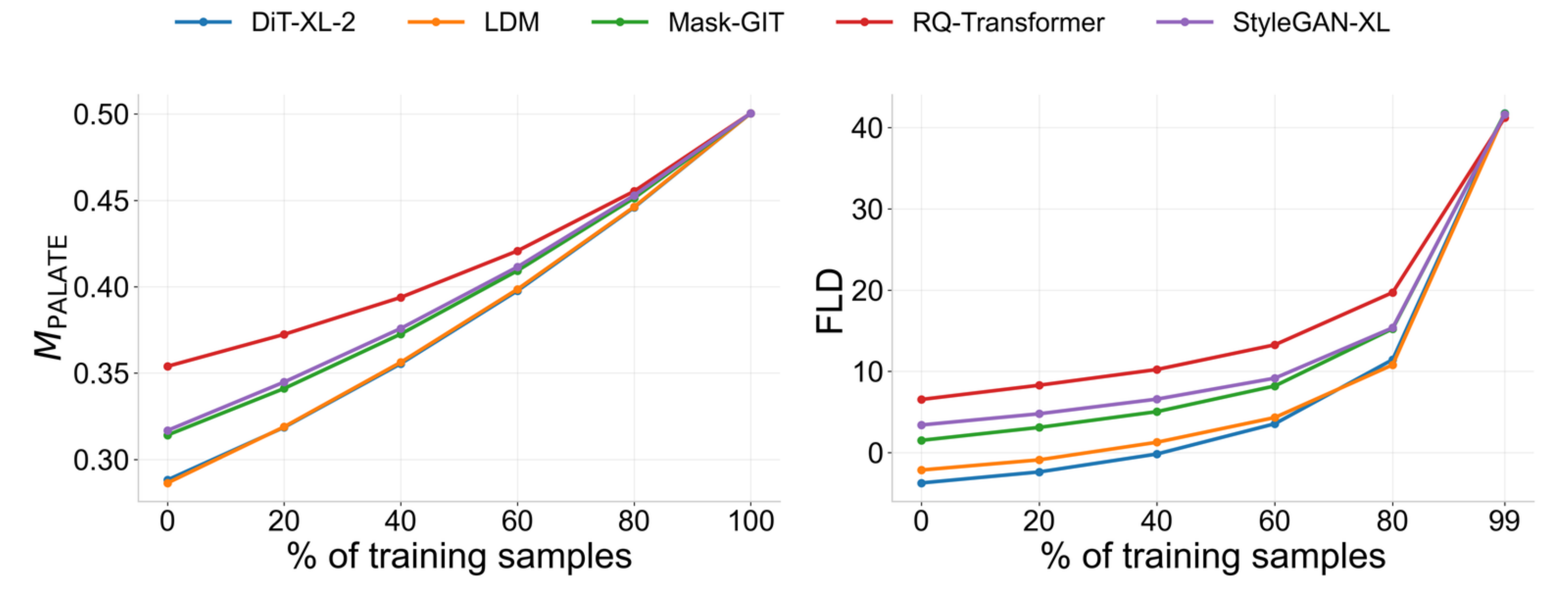}
\end{subfigure}

\caption{Performance of $M_{\text{PALATE}}$ (our) and FLD on mixtures of generated and training images, with the proportion of training samples varying from $0\%$ to $100\%$. FLD becomes numerically unstable at $100\%$ (reaching values up to $\sim10^{7}$) and is therefore shown only up to $99\%$. {\em Top}: CIFAR-10. {\em Bottom}: ImageNet.}
\label{fig:metrics_interpolation}
\end{figure}


\begin{figure}[t]
\centering
\includegraphics[width=0.55\linewidth]{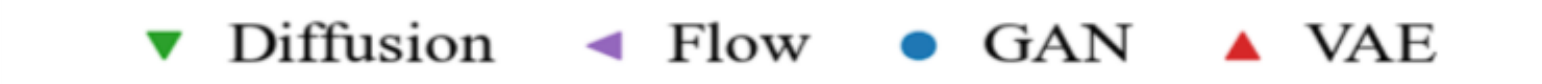}\\[1mm]
\includegraphics[width=0.325\linewidth]{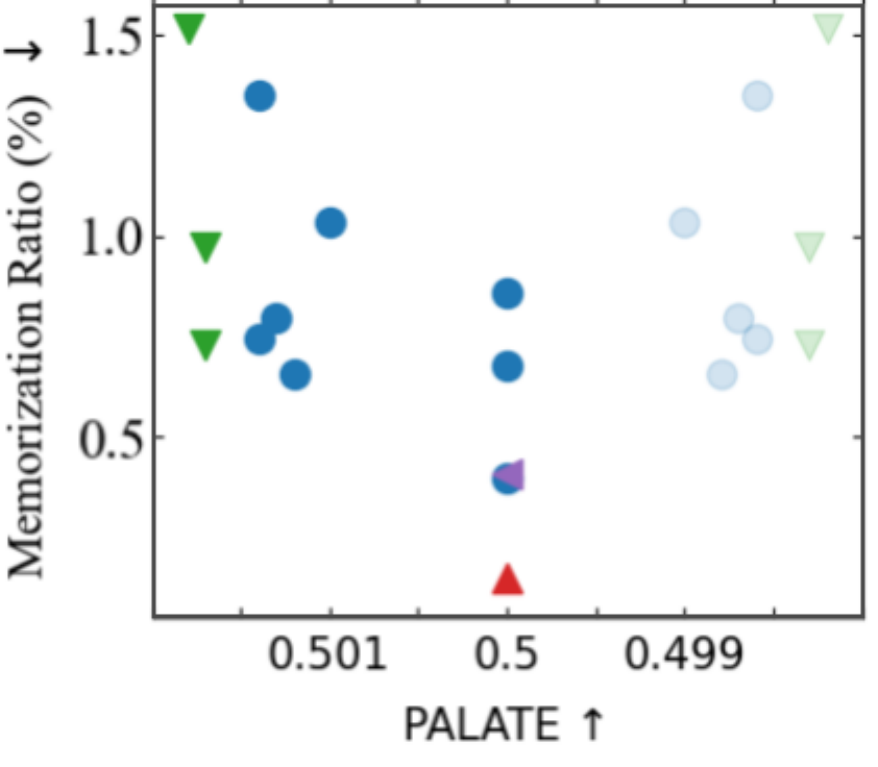}\hfill
\includegraphics[width=0.27\linewidth]{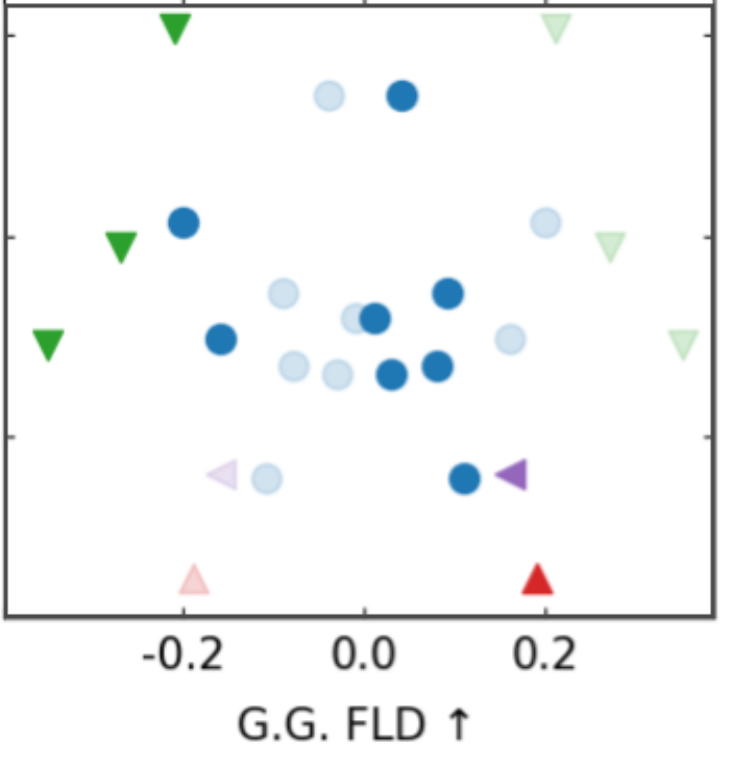}\hfill
\includegraphics[width=0.263\linewidth]{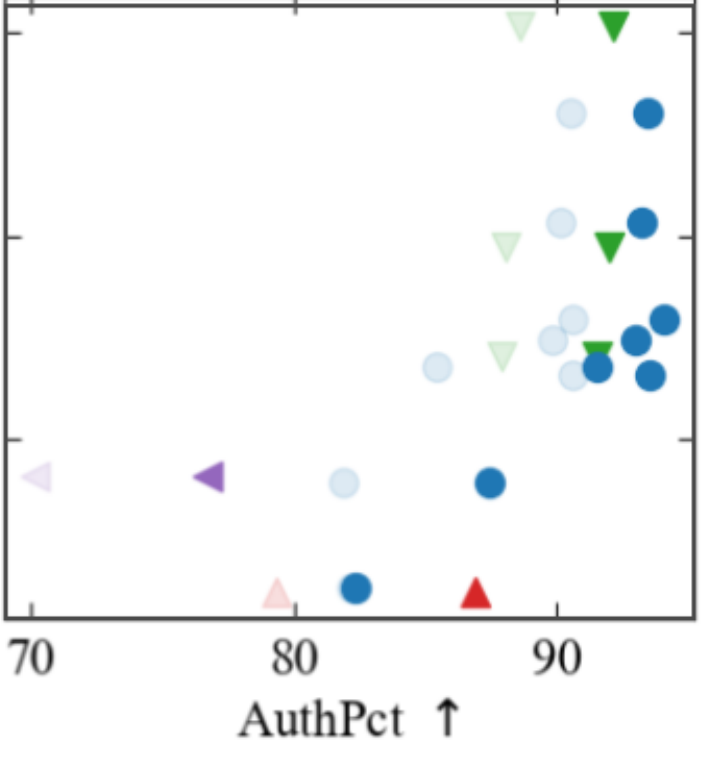}\\[1mm]
\includegraphics[width=0.27\linewidth]{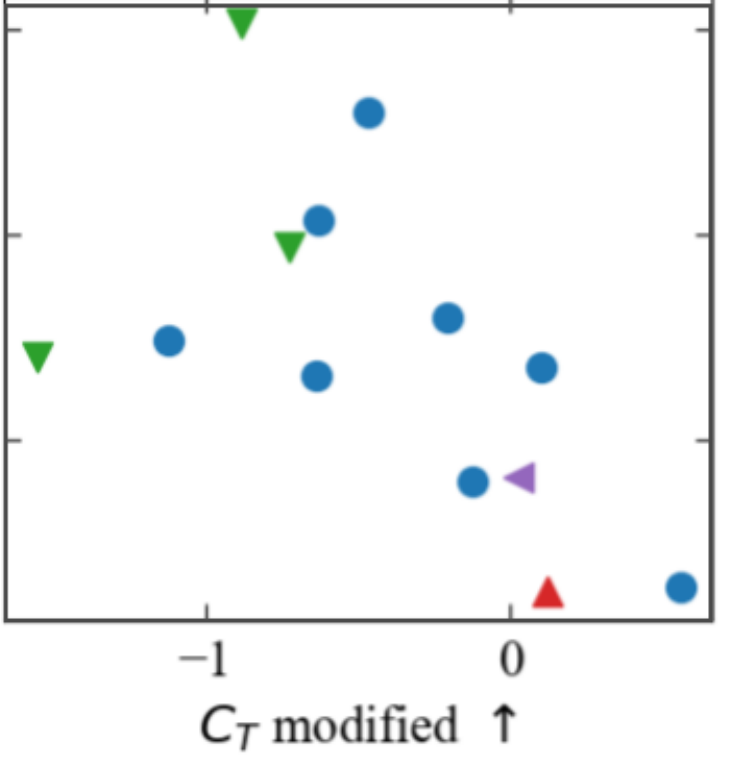}\hspace{4mm}
\includegraphics[width=0.268\linewidth]{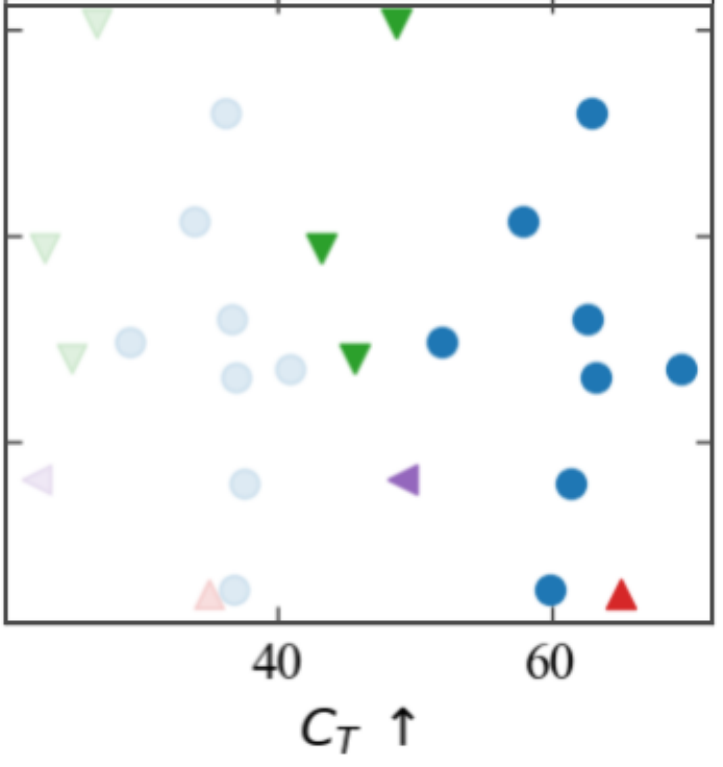}
\caption{Memorization ratios vs. metric values for CIFAR-10-trained models. Swapped train/test memorization metrics are shown with low opacity. Results for $C_T$, $C_T$ mod., and AuthPct are taken from \cite{stein2024exposing}. The $x$-axis of \our{} (our) is inverted.
}

\end{figure}

\begin{figure}[t]
\centering

\begin{subfigure}[t]{0.92\textwidth}
    \centering
    \includegraphics[width=\linewidth]{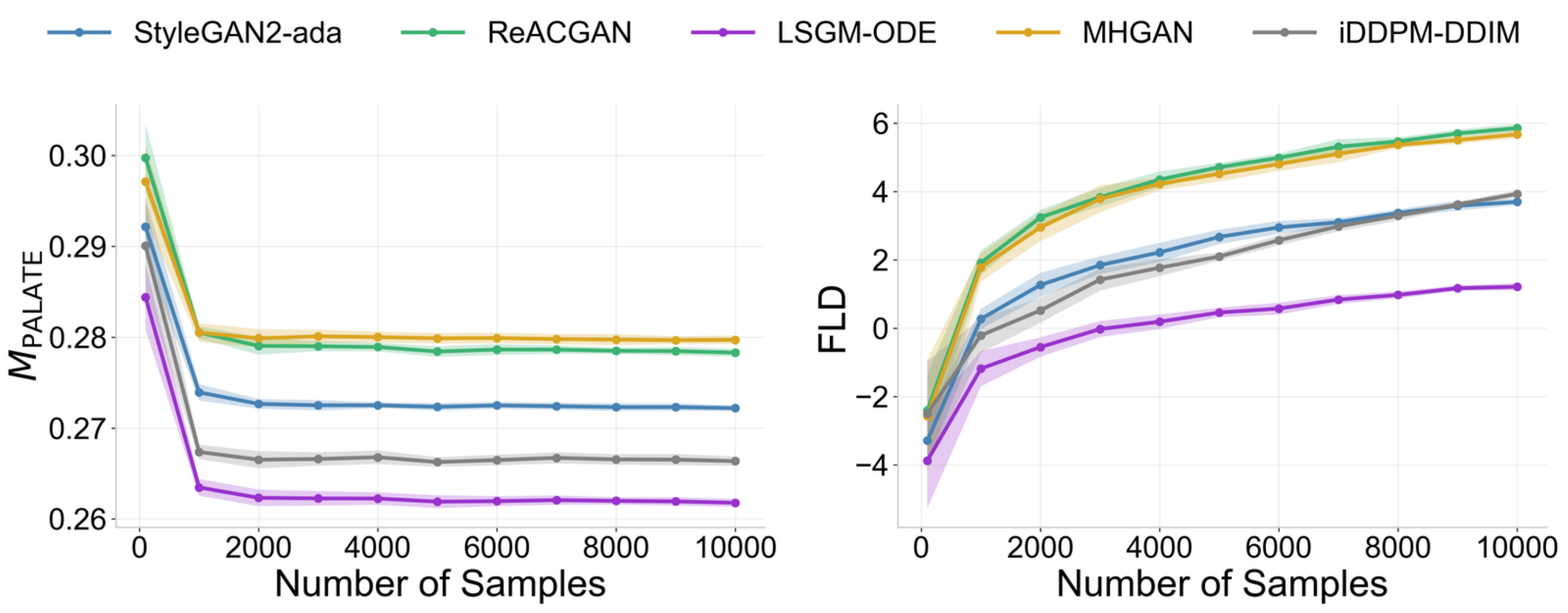}
\end{subfigure}\\[5mm]
\begin{subfigure}[t]{0.92\textwidth}
    \centering
    \includegraphics[width=\linewidth]{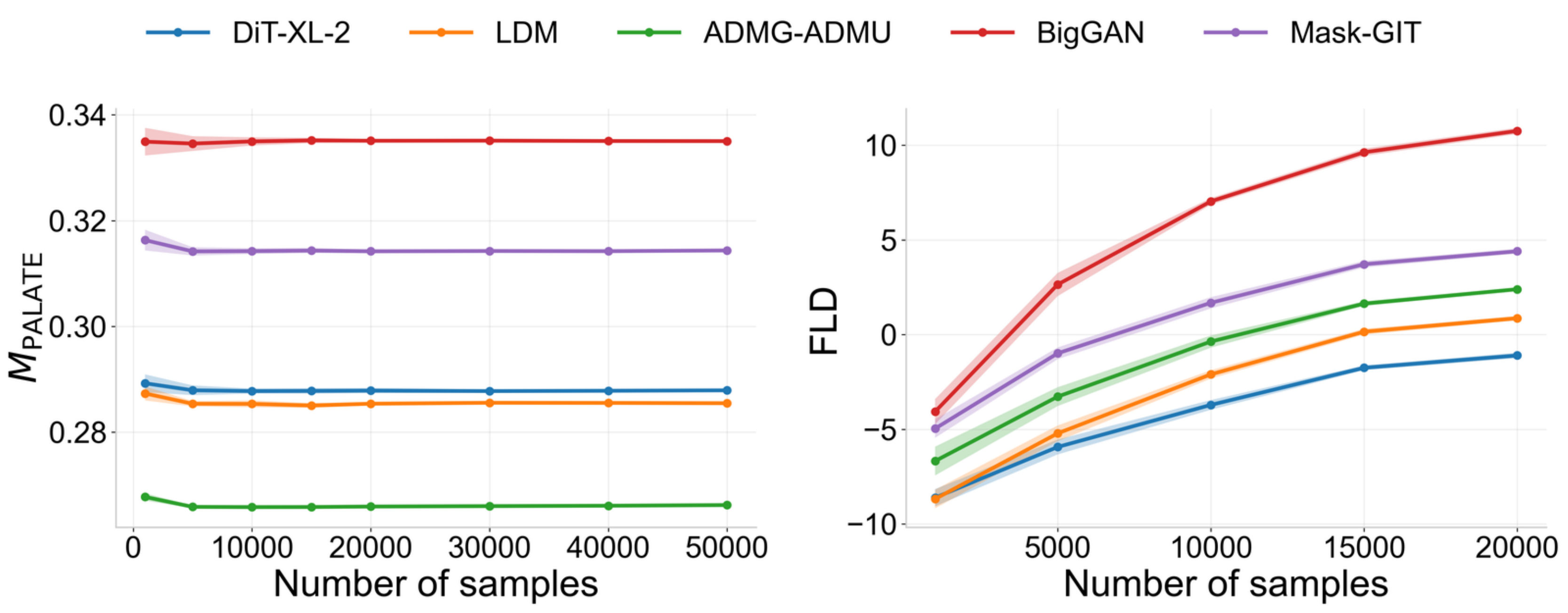}
\end{subfigure}

\caption{Stability of $M_{\text{PALATE}}$ (our) and FLD as a function of sample size (mean $\pm$ standard deviation over 10 random seeds). {\em Top:} CIFAR-10. {\em Bottom:} ImageNet.}
\label{fig:stability}
\end{figure}

For fidelity assessment on CIFAR-10, most metrics exhibit strong positive correlations with Human Error Rate, with $\rho$ ranging from 0.88 (CMMD) to 0.98 ($M_\text{PALATE}$). Precision and Density are notable exceptions, showing negative correlations. On ImageNet, correlations remain consistently positive but are weaker overall, ranging from 0.63 (CMMD) to 0.78 (KD).

For diversity assessment, where no explicit reference is available, most metrics on CIFAR-10 are strongly correlated with each other. Mean per-class Vendi Score is an exception, showing only weak positive correlations ($\rho\in[0.17,0.23]$). Several metric pairs reach near-perfect agreement (up to $\rho=0.99$ for $M_\text{\our{}}$--FD--KD, FLD--Recall, and FD--Coverage). On ImageNet, moderate-to-strong positive correlations ($\rho\geq0.68$) are largely preserved, except for Recall, which shows a negative correlation with mean per-class Vendi Score.

Memorization assessment is more challenging. Most metrics on CIFAR-10 show weak correlations with Memorization Ratio. The negative correlations observed for holistic metrics ($M_\text{\our{}}$ and FLD) are expected, since memorization assessment in these methods serves only as a refinement of the fidelity--diversity trade-off. Other memorization-related metrics, including $C_T$, $C_T$ mod., and AuthPct, show limited or inconsistent correlations\footnote{This is consistent with the conclusions reported in \cite[Sec.~5.2]{stein2024exposing}.
}. In contrast, \our{} ($\rho=0.65$) and Generalization Gap FLD ($\rho=0.63$) exhibit moderate positive correlations. On ImageNet, where Memorization Ratios are unavailable\footnote{Due to the substantial memory requirements of ImageNet and prior findings indicating that memorization is not a primary concern for this dataset~\citep{stein2024exposing}, we do not compute the Memorization Ratio for ImageNet.}, all memorization-oriented metrics exhibit moderate-to-strong positive correlations, with the highest being $\rho = 0.95$ between $C_T$ and Generalization Gap FLD.

Overall, these results indicate that $M_\text{\our{}}$ provides a reliable single-score holistic evaluation framework, while \our{} offers a more direct and consistent assessment of memorization. Both metrics perform competitively with existing baselines across the considered real-world datasets.

\paragraph{Sample Fidelity}
\cref{fig:fidelity} illustrates the behavior of $M_\text{\our{}}$ and FLD under increasing image distortions applied to samples generated by PFGM++ trained on CIFAR-10. For mild, largely imperceptible transformations (Posterize, Light Blur, JPG90), both metrics degrade slightly relative to the baseline, yet still rate the samples higher than those from StyleGAN2-ada. In contrast, stronger distortions, particularly Elastic Transform and Heavy Blur, lead to a pronounced drop in both metrics, as expected. Overall, the degradation in metric values is well aligned with the severity of the applied perturbations.

\paragraph{Sample Diversity}

We evaluate sensitivity to sample diversity by varying the number of semantic classes in generated samples from CIFAR-10 and ImageNet; results appear in \cref{fig:sample_diverity}. When progressively increasing the number of classes while fixing the total sample size (via balanced data replication), $M_\text{\our{}}$ decreases monotonically, matching the behavior of FLD and correctly capturing the rising diversity. In \ref{sec:add_exp}, we additionally report Recall and Coverage for the same experiment, which exhibit similar monotonic behavior as diversity increases.

To test robustness further, we introduce controlled class imbalance by duplicating samples from a single class while leaving others unchanged. As shown in \cref{fig:class_imbalance}, $M_\text{\our{}}$ consistently increases with the duplication factor, detecting diversity loss from redundancy and mode concentration. In contrast, FLD remains largely insensitive to this manipulation. These results demonstrate that $M_\text{\our{}}$ responds to both emerging semantic modes and diversity degradation from frequency collapse.

\paragraph{Sample Novelty}
To assess novelty, we progressively mix generated samples with samples from the training set for models trained on CIFAR-10 and ImageNet. As shown in \cref{fig:metrics_interpolation}, both $M_\text{PALATE}$ and FLD increase as memorization grows; however, $M_\text{PALATE}$ exhibits a smooth and gradual rise, whereas FLD remains nearly flat for most of the interpolation range and then undergoes an abrupt transition near full memorization. Notably, FLD becomes numerically unstable at $100\%$ training data, growing by several orders of magnitude and saturating the plot scale, while $M_\text{PALATE}$ remains stable and interpretable throughout.

Additionally, in Figure~7, we extend the experiment proposed in \cite{stein2024exposing}, which examines whether memorization-oriented metrics on CIFAR-10 can practically distinguish models based on their Memorization Ratios. All metrics except AuthPct exhibit the correct trend, indicating their ability to differentiate models with varying degrees of memorization, which is consistent with the Spearman correlations shown in \cref{fig:metric-correlations}. However, when the training set is replaced by a test set that the model cannot memorize, all metrics except PALATE and Generalization Gap FLD produce nearly identical trends, suggesting that they are largely driven by other factors such as fidelity or mode collapse rather than memorization \cite{stein2024exposing}.

\begin{figure}[t] 
    \centering
    \includegraphics[width=1\columnwidth]{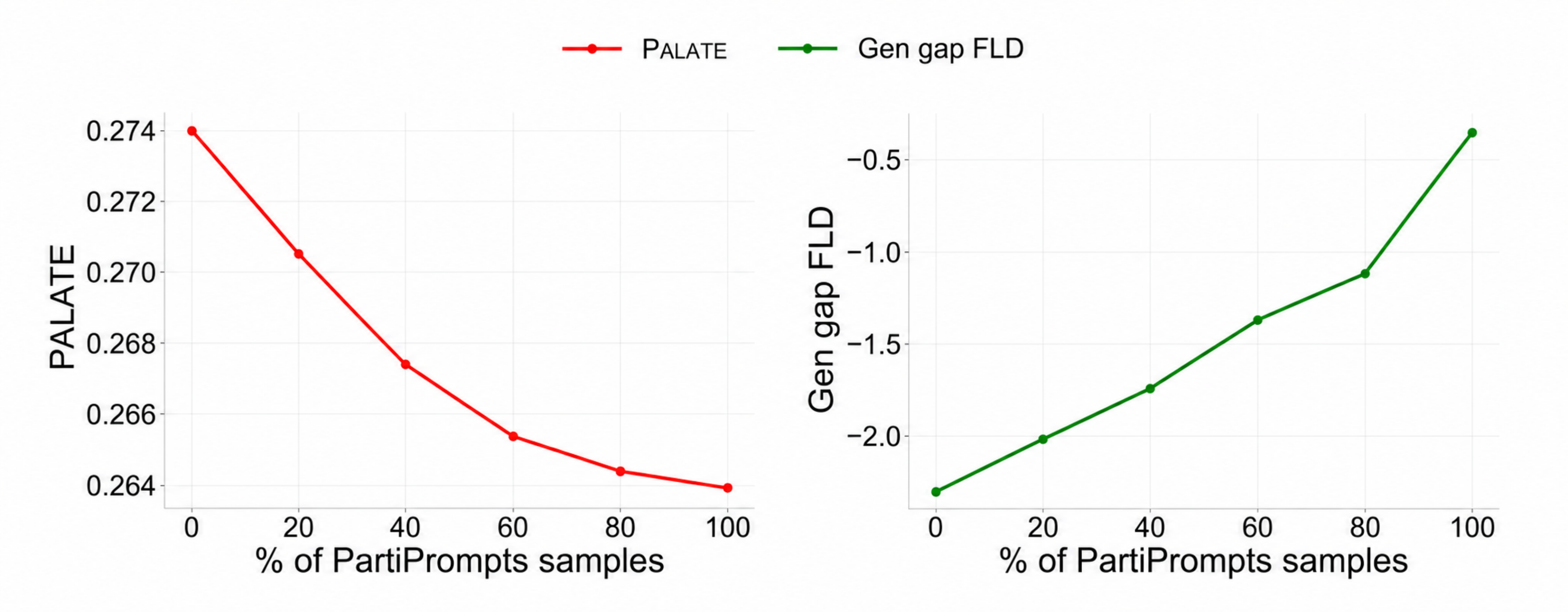}
    \caption{
    Performance of PALATE (our) and Generalization Gap FLD on prompts interpolating between MS-COCO training captions and PartiPrompts, with the proportion of PartiPrompts varying from $0\%$ to $100\%$.}\vskip-5mm
  \label{fig:parti}
\end{figure}

\begin{figure}[t] 
    \centering
    \includegraphics[width=0.9\columnwidth]{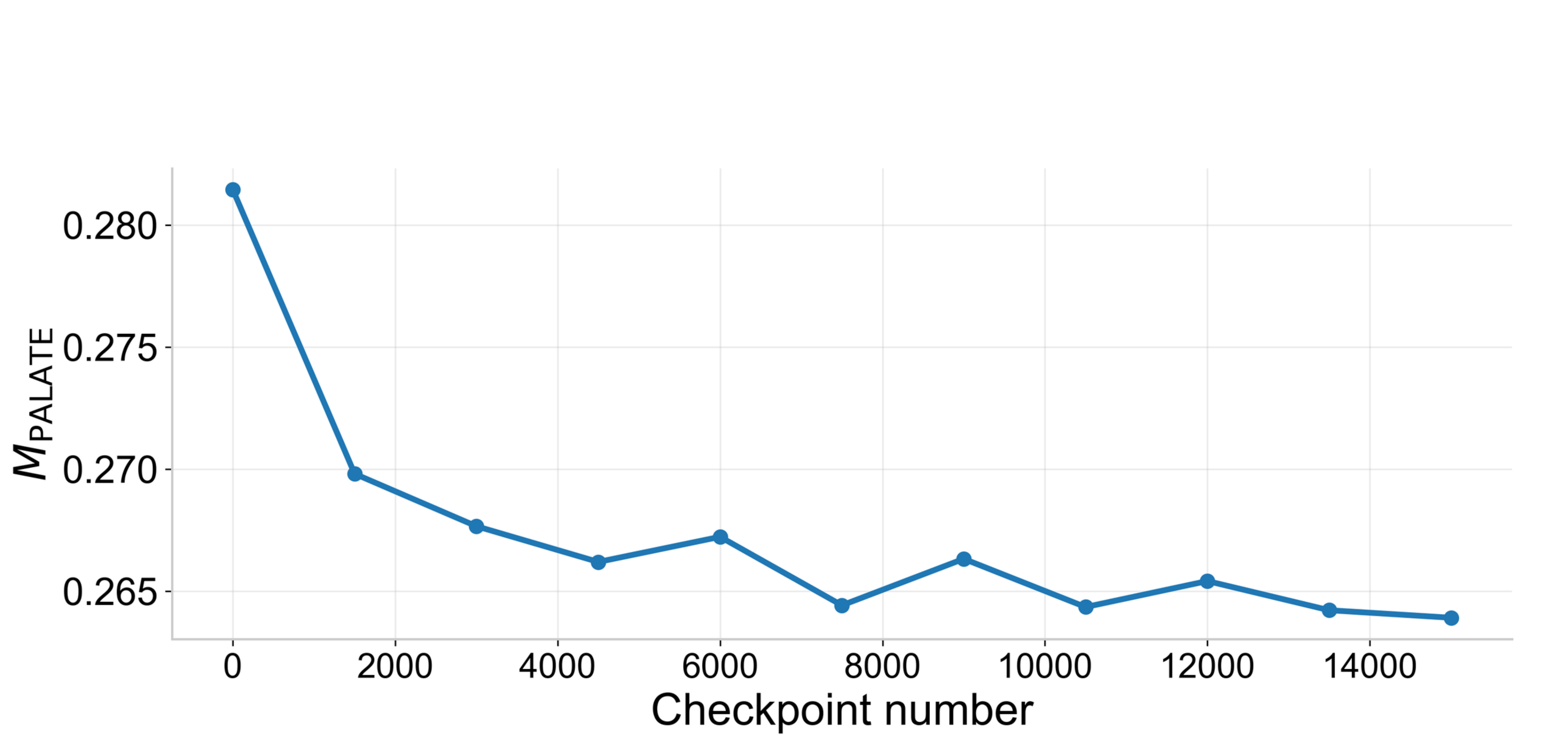}
    \caption{
    Evolution of $M_{\text{PALATE}}$ (our) across LoRA fine-tuning checkpoints of Stable Diffusion v1.5 on $5{,}000$ MS-COCO caption--image pairs.}

    \label{fig:text-to-image_fidelity}
\end{figure}

\begin{figure}[t] 
    \centering
    \includegraphics[width=0.9\columnwidth]{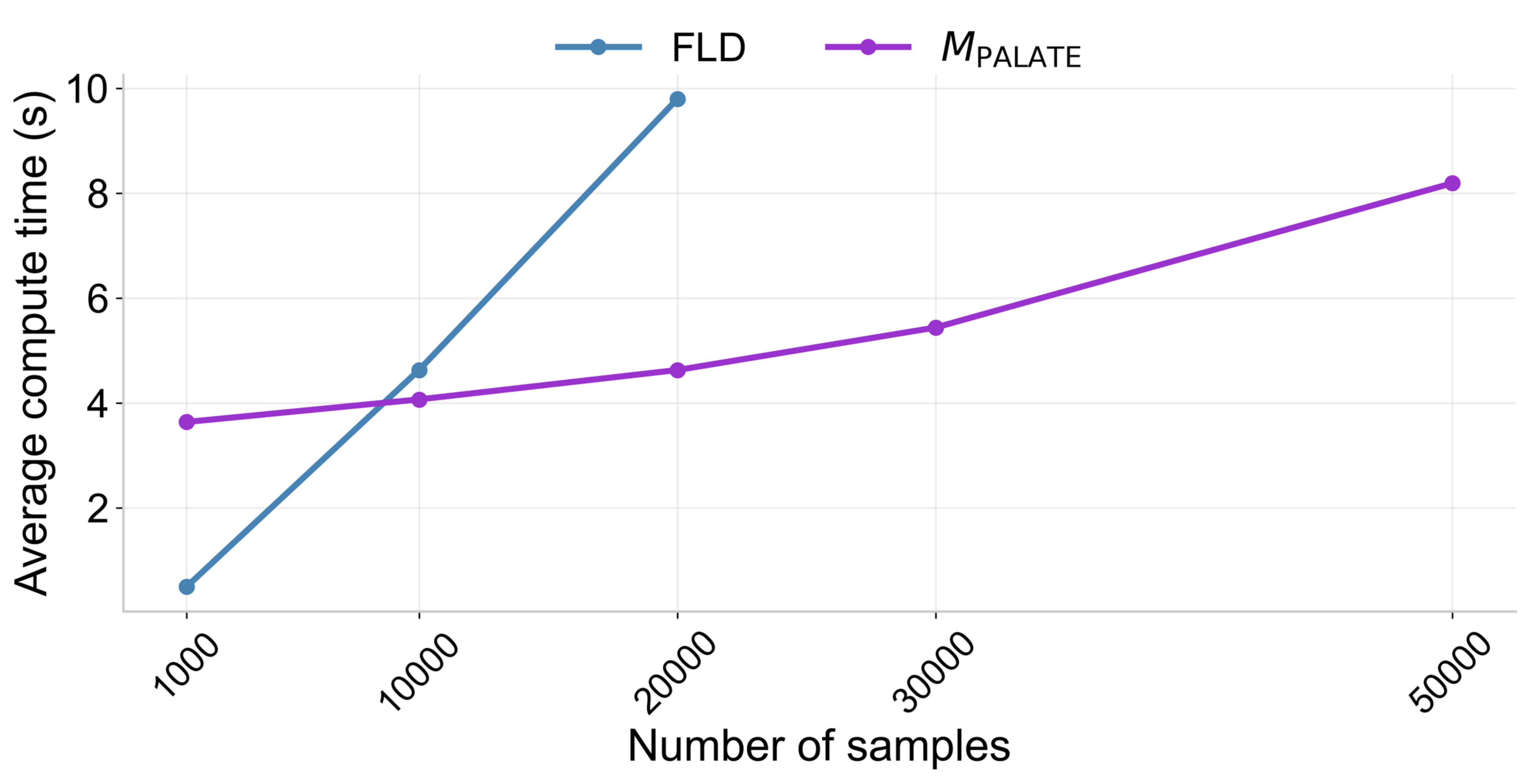}
    \caption{Computation time of $M_{\text{PALATE}}$ (our) and FLD on ImageNet as a function of sample size. FLD is shown only up to $20{,}000$ samples due to memory limitations at larger scales. All runtimes are averaged over 10 random seeds.}

    \label{fig:compute}
\end{figure}

\paragraph{Text-to-Image Overfitting Setup}
To assess semantic overfitting in text-to-image generation, we fine-tune Stable Diffusion v1.5 on $5{,}000$ MS-COCO training caption–image pairs using LoRA, and evaluate the resulting model on prompts that progressively shift away from the training distribution. Specifically, we interpolate between MS-COCO training captions and PartiPrompts ($0\%$ to $100\%$ PartiPrompts) and generate samples using identical sampling hyperparameters and a fixed random seed.

As shown in \cref{fig:parti}, generations conditioned on training captions exhibit the strongest memorization signal, reflected by higher $\text{\our{}}$ and a lower Generalization Gap FLD. As prompts shift toward PartiPrompts~\citep{yu2022scaling}, $\text{\our{}}$ slightly decreases while the Generalization Gap FLD slightly increases. This effect is expected to be mild, since the model is not directly evaluated on training images, but rather on new generations conditioned on training captions, which only indirectly encourages memorization.

Additionally, we evaluate $M_\text{PALATE}$ across checkpoints thro\-ughout fine-tuning. As shown in \cref{fig:text-to-image_fidelity}, we observe that $M_\text{PALATE}$ decreases over training, which is consistent with the model improving fidelity rather than exhibiting strong image-level memorization. We hypothesize that in this regime (LoRA fine-tuning on only $5{,}000$ examples), training primarily refines the pretrained model’s caption–image alignment and overall generation quality, while the limited number of updated parameters makes exact training-image reproduction less likely. More details on our Stable Diffusion fine-tuning are provided in \ref{app:impdet}.

\paragraph{Computational Efficiency and Stability}
We compare $M_\text{PALATE}$ and FLD under varying sample sizes on CIFAR-10 and ImageNet. As shown in \cref{fig:stability}, $M_\text{PALATE}$ stabilizes quickly, producing consistent values and preserving model rankings even for small sample sizes. In contrast, FLD remains sensitive to sample size and continues to fluctuate in this regime.

These stability differences are reflected in computational efficiency. As shown in \cref{fig:compute}, $M_\text{PALATE}$ is consistently faster than FLD beyond $10{,}000$ samples on ImageNet. Moreover, FLD incurs substantial memory overhead and cannot be computed for $30{,}000$ samples or more on an RTX~4090, despite requiring larger sample sizes for stable estimates. The efficiency of $M_\text{PALATE}$ stems largely from its reliance on matrix multiplications, which are highly optimized in modern deep learning frameworks.

\section{Conclusions}\label{sec:conclusions}
We introduced \our{}, a novel memorization-aware enhancement to the evaluation of deep generative models, grounded in a peculiar application of the law of total expectation. The proposed formulation enables sensitivity to sample memorization and overfitting, which are not explicitly captured by many existing evaluation metrics. By integrating \our{} with an MMD-based baseline and leveraging DINOv3 feature representations, we obtain a holistic single-score evaluation metric that jointly captures fidelity, diversity, and novelty of generated samples while remaining computationally efficient. Extensive experiments on CIFAR-10 and ImageNet demonstrate that our approach reduces computational overhead while achieving evaluation performance that is comparable to, and in several cases surpasses, state-of-the-art alternatives.

\paragraph{Limitations}\label{paragraph:Limitations}
At present, our evaluation is limited to deep generative models trained on image datasets, along with a single text-to-image overfitting scenario. Extending the proposed framework to additional data modalities, model families, and training regimes constitutes an important direction for future work.

\appendix

\makeatletter
\@addtoreset{figure}{section}%
\@addtoreset{table}{section}%
\makeatother
\setcounter{figure}{0}
\setcounter{table}{0}

\section{Deriving the Test-Region Mass}
\label{sec:proofs}

Let $g\colon\mathcal{X}\to\mathbb{R}$ be any nontrivial measurable function. Applying the law of total expectation to $g(X)$ and the partition
\begin{equation}
\Omega=\{\omega:X(\omega)\in\mathcal{M}_\text{test}\}\cup\{\omega:X(\omega)\in\mathcal{M}_\text{train}\},
\end{equation}
we obtain
\begin{equation}\label{eq:loteg}
\E(g(X))
=
a\,\E(g(X)\mid X\in\mathcal{M}_\text{test})
+(1-a)\,\E(g(X)\mid X\in\mathcal{M}_\text{train}).
\end{equation}

Replacing expectations with empirical averages yields
\begin{equation}\label{eq:loteeq}
\frac{1}{m+n}
\Big(
\sum_{i=1}^m g(x_i^\text{train})+\sum_{i=1}^n g(x_i^\text{test})
\Big)
=
\frac{1-a}{m}\sum_{i=1}^m g(x_i^\text{train})
+
\frac{a}{n}\sum_{i=1}^n g(x_i^\text{test})
\end{equation}
Solving for $a$ gives
\begin{equation}\label{eq:lotetrans}
a=\frac{n}{m+n},
\end{equation}
which depends only on the relative sizes of the training and test datasets and is independent of the specific data values.

\section{Additional Experimental Results}
\label{sec:add_exp} 
\begin{figure}[]
\centering

\begin{subfigure}[t]{0.75\textwidth}
    \centering
    \includegraphics[width=\linewidth]{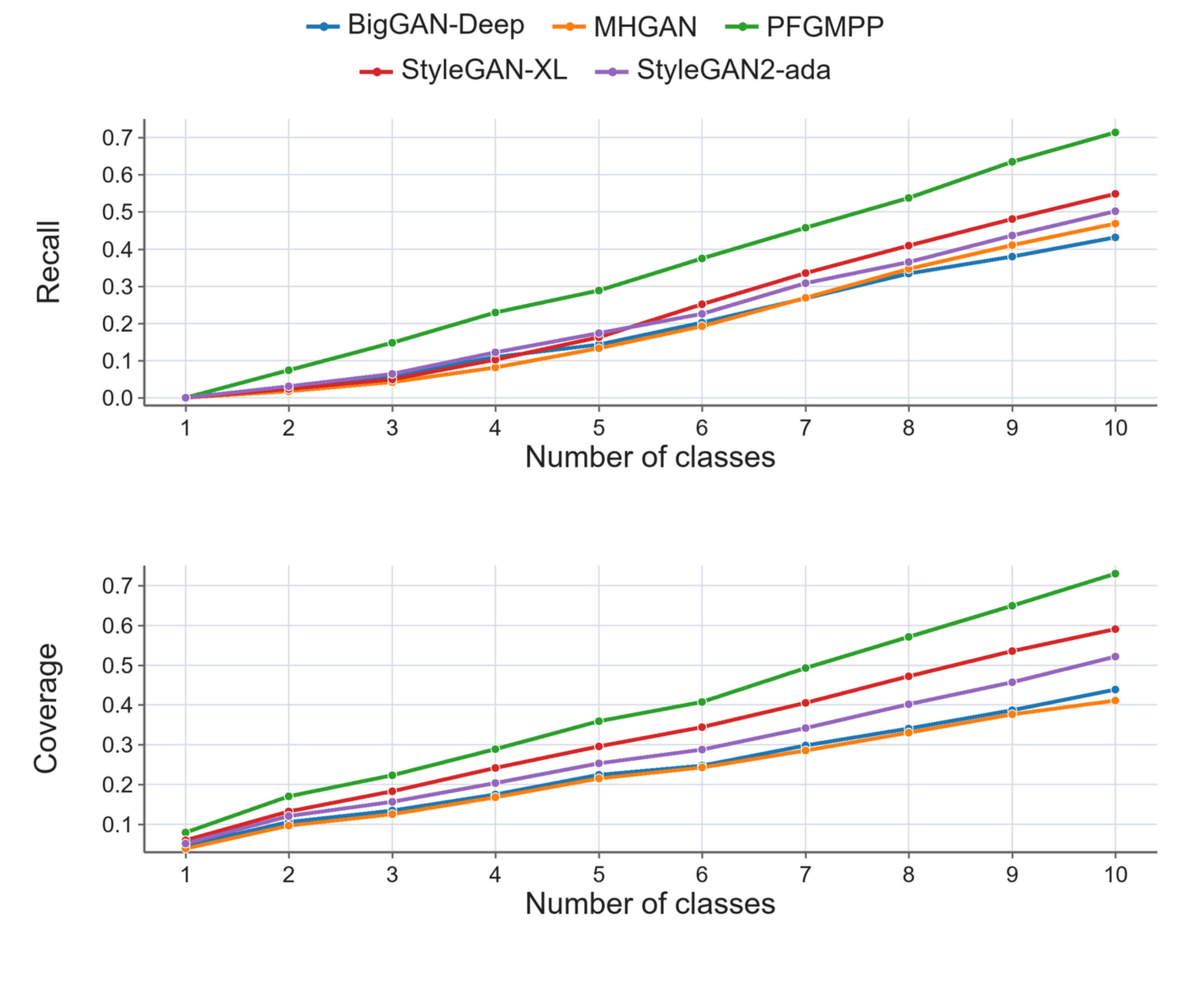}
\end{subfigure}\\[3mm]

\begin{subfigure}[t]{0.75\textwidth}
    \centering
    \includegraphics[width=\linewidth]{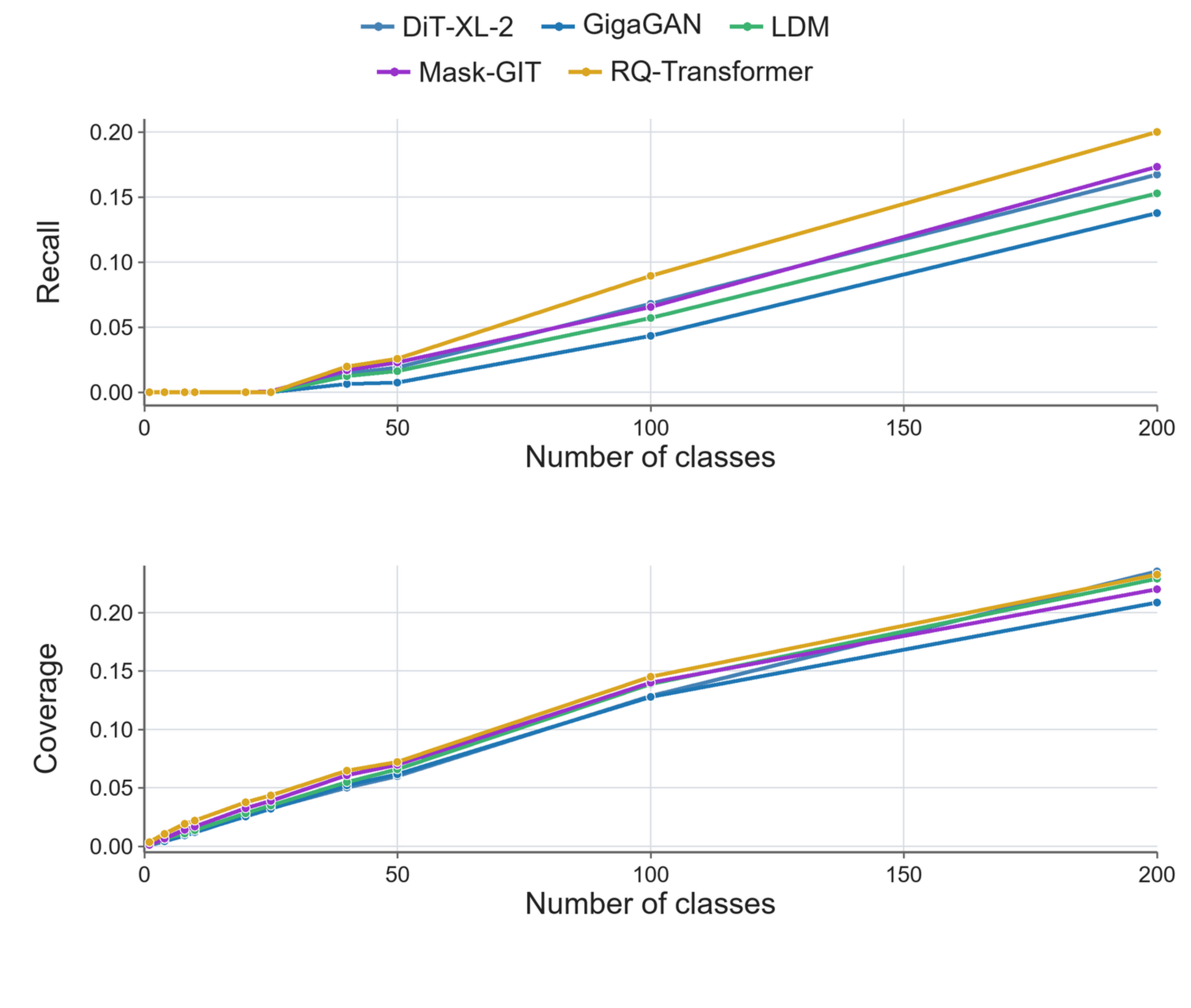}
\end{subfigure}

\caption{Capability of Recall and Coverage metrics to capture increasing sample diversity as the number of classes grows, while keeping the total number of generated samples fixed at $10{,}000$.
{\em Top}: CIFAR-10.
{\em Bottom}: ImageNet.}
\label{fig:recallcoverage}
\end{figure}

\begin{figure}[t] 
    \centering
    \includegraphics[width=0.9\columnwidth]{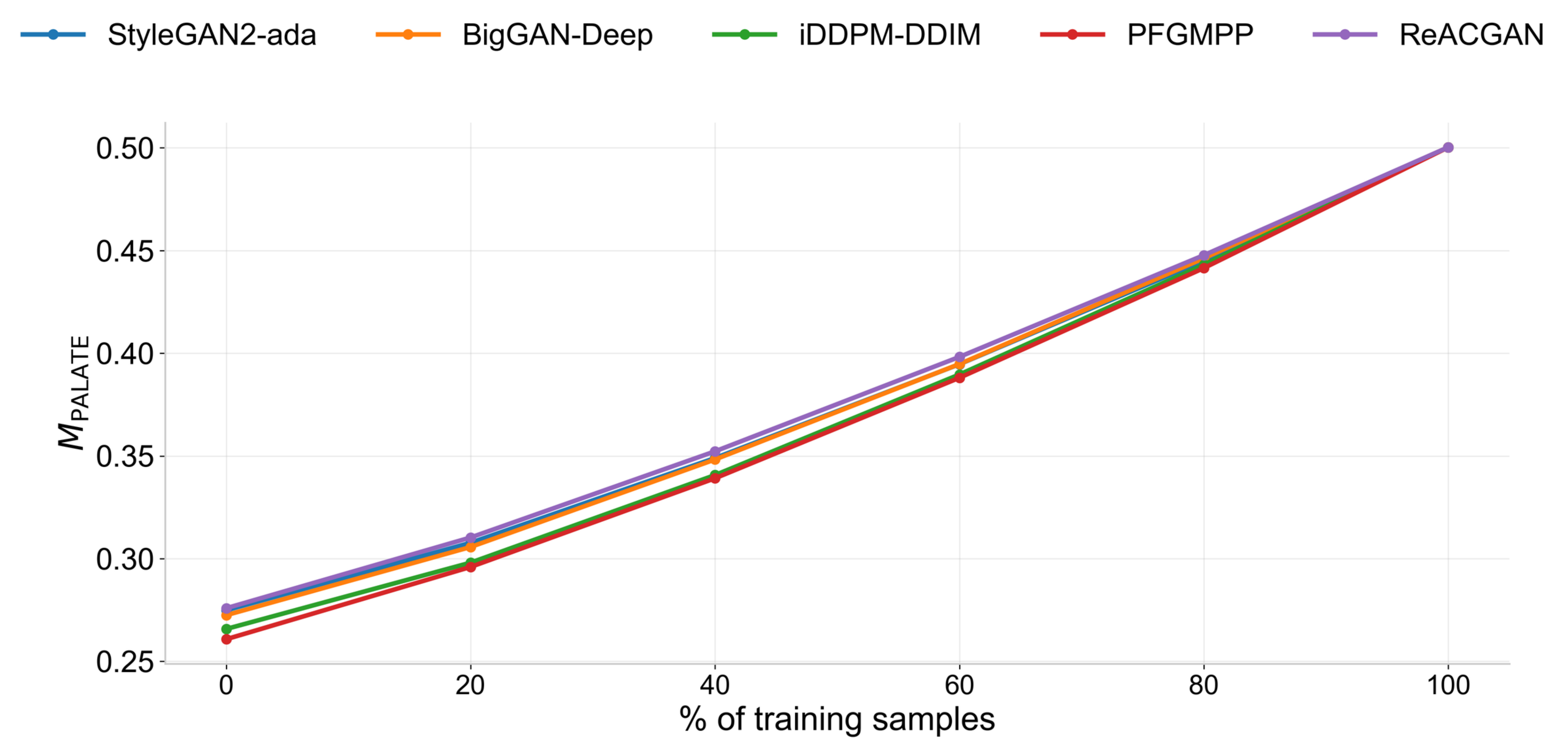}
    \caption{Performance of $M_{\text{PALATE}}$ (our) computed on DINOv2 embeddings on mixtures of generated and training images, with the proportion of training samples varying from $0\%$ to $100\%$ on CIFAR-10.}
    \label{fig:inter_dinov2}
\end{figure}

\begin{figure}[t] 
    \centering
    \includegraphics[width=0.92\columnwidth]{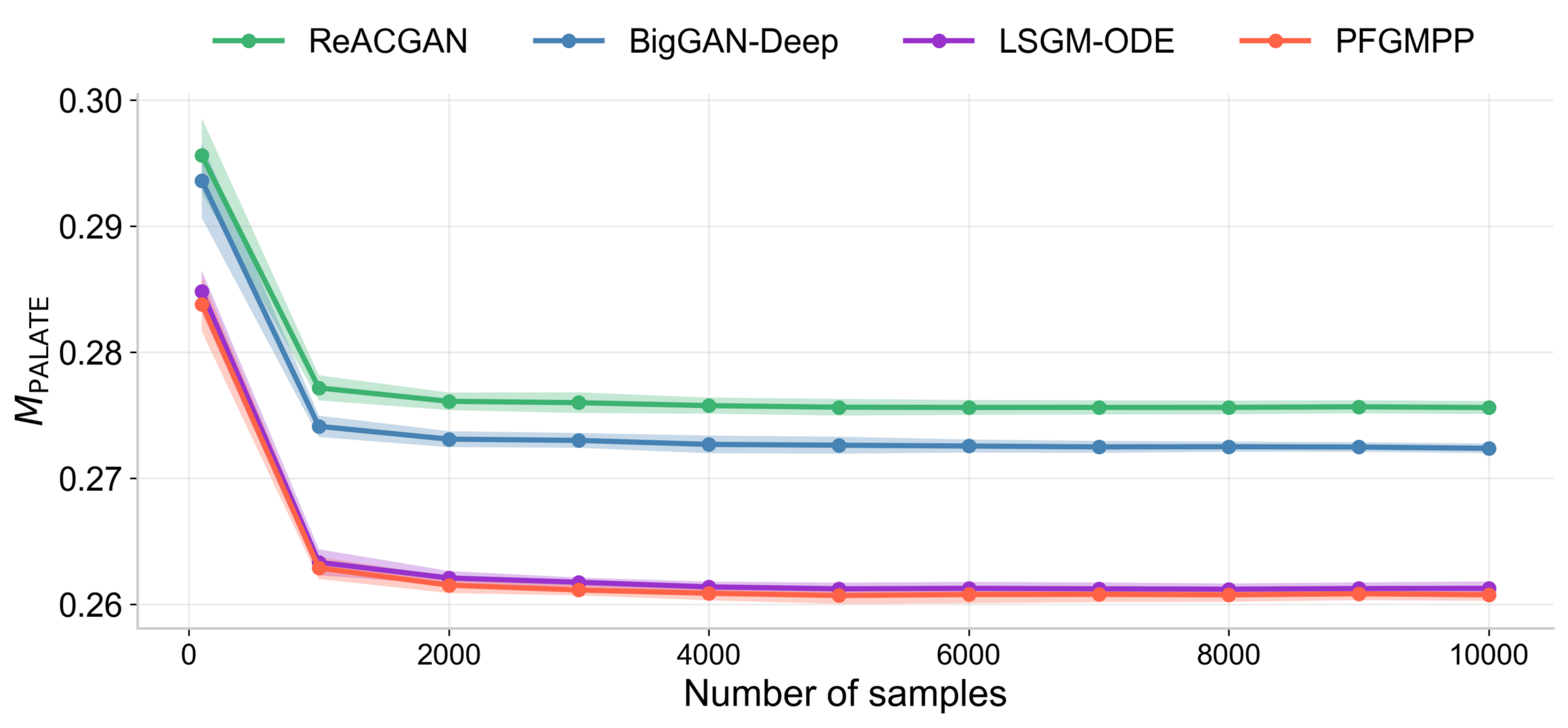}
    \caption{Stability of $M_{\text{PALATE}}$ (our) computed on DINOv2 embeddings as a function of sample size (mean $\pm$ standard deviation over 10 random seeds) on CIFAR-10.}

\label{fig:stab_dinov2}
\end{figure}

\begin{figure}[t] 
    \centering
    \includegraphics[width=.9\columnwidth]{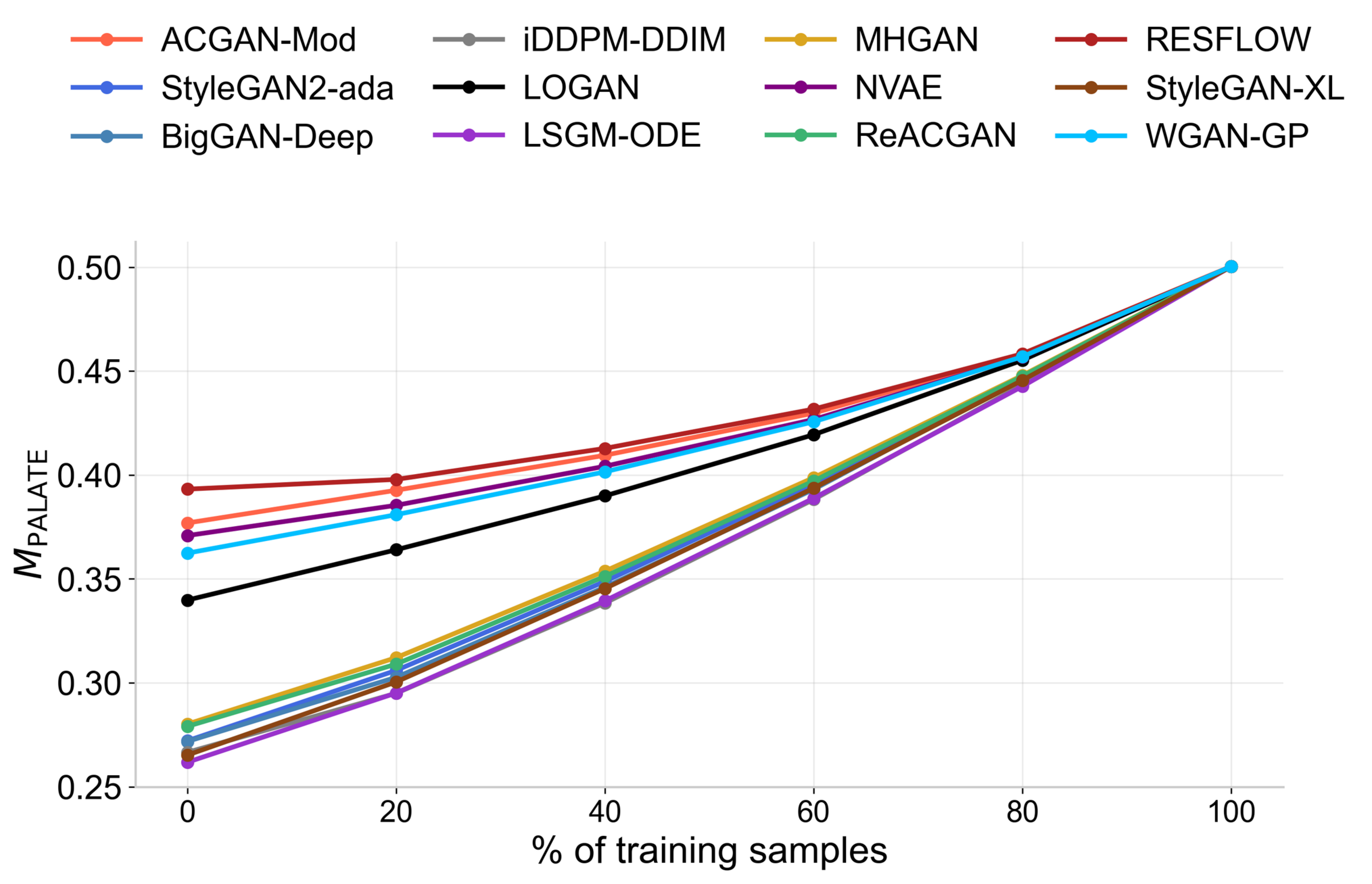}
    \caption{Performance of $M_{\text{PALATE}}$ (our) computed on DINOv3 embeddings on mixtures of generated and training images, with the proportion of training samples varying from $0\%$ to $100\%$ on CIFAR-10.}

    \label{fig:inter_dinov3}
\end{figure}

The results of our additional experiments are provided in Figures \ref{fig:recallcoverage}, 
\ref{fig:inter_dinov2},
 \ref{fig:stab_dinov2}
 and \ref{fig:inter_dinov3}.
\suppressfloats[t]

\section{Implementation Details}\label{app:impdet}

\begingroup
\setlength{\emergencystretch}{1em}
\paragraph{Metrics Computation}
We provide explicit formulas for $\text{PALATE}$ and $M_\text{PALATE}$
metrics based on real and generated samples
\par
\endgroup
\[
\begin{gathered}
{\rm x}_{\text{train}}=\{x^\text{train}_1,\ldots,x^\text{train}_n\},\\
{\rm x}_{\text{test}}=\{x^\text{test}_1,\ldots,x^\text{test}_n\},
\quad\text{and}\quad {\rm y}=\{y_1,\ldots,y_n\}.
\end{gathered}
\]
While $M_\text{PALATE}$ is computed using all generated samples,
$\text{PALATE}$ score operates only on the subset
\[
{\rm y}_{\text{data}}=\{y^\text{data}_1,\ldots,y^\text{data}_k\}\subset{\rm y}
\]
assigned to the data manifold via the KDE-based filter.

Let $\bar{k}_{A,B}$ denote the $V$-statistic kernel average between feature
sets $\phi(A)$ and $\phi(B)$. Define
\begin{align}
D_{\text{test}}
&=
\bar{k}_{{\rm x}_\text{test},{\rm x}_\text{test}}
+
\bar{k}_{{\rm y}_\text{data},{\rm y}_\text{data}}
-
2\bar{k}_{{\rm x}_\text{test},{\rm y}_\text{data}},
\\
D_{\text{train}}
&=
\bar{k}_{{\rm x}_\text{train},{\rm x}_\text{train}}
+
\bar{k}_{{\rm y}_\text{data},{\rm y}_\text{data}}
-
2\bar{k}_{{\rm x}_\text{train},{\rm y}_\text{data}}.
\end{align}
Then
\begin{equation}
\text{PALATE}
=
\frac{D_{\text{test}}}
{D_{\text{test}}+D_{\text{train}}}.
\end{equation}
Similarly, the scaled DMMD baseline is
\begin{equation}
\text{SCALE}(\text{DMMD})
=
\frac{
\bar{k}_{{\rm x}_\text{test},{\rm x}_\text{test}}
+
\bar{k}_{{\rm y},{\rm y}}
-
2\bar{k}_{{\rm x}_\text{test},{\rm y}}
}{
\bar{k}_{{\rm x}_\text{test},{\rm x}_\text{test}}
+
\bar{k}_{{\rm y},{\rm y}}
}.
\end{equation}
The final holistic metric is
\begin{equation}
M_\text{PALATE}
=
\frac{1}{2}\,\text{SCALE}(\text{DMMD})
+
\frac{1}{2}\,\text{PALATE}.
\end{equation}
Here, $\phi$ denotes the DINOv3 feature extractor. For example,
\begin{equation}
\bar{k}_{{\rm x}_\text{test},{\rm x}_\text{test}}
=
\frac{1}{n^2}
\sum_{i,j=1}^n
\exp\!\left(
-\frac{
\|\phi(x_i^\text{test})-\phi(x_j^\text{test})\|^2
}{2\sigma^2}
\right),
\end{equation}
with analogous expressions for the remaining train, test, and generated
sample terms.

\paragraph{Bandwidth Selection}
The kernel bandwidth $\sigma$ is tuned to ensure sensitivity to memorization while preserving fidelity and diversity information in $M_\text{PALATE}$.
We follow the median heuristic of \cite{gretton2012kernel}.
Specifically, for each dataset we search over
$\{\sqrt{\text{Med}}/k \mid k=1,2,3\}$,
where $\text{Med}$ denotes the median of squared pairwise Euclidean distances between features of all real samples. In all experiments, we set $k=2$, i.e., $\sigma=\sqrt{\text{Med}}/2$.

To emphasize memorization effects, the bandwidth used in the $\text{PALATE}$ term is chosen to be three times smaller than that used in $\text{SCALE}(\text{DMMD})$.
This conservative choice increases sensitivity to subtle differences between training and test samples, which are assumed to be drawn from the same underlying data distribution $p_X$.
The selected bandwidth values are summarized in \cref{tab:sigma}.

\begin{table}[h]
    \centering
    \caption{Bandwidth $\sigma$ values used in our experiments.}
    \begin{tabular}{ccc}
     \toprule
      & \text{SCALE(DMMD)} & \text{PALATE} \\
     \midrule
     CIFAR-10  & 10.5 & 3.5 \\
     ImageNet  & 11.6   & 3.9  \\
     MS-COCO   & 10.9   & 3.6  \\
     \bottomrule
    \end{tabular}
    \label{tab:sigma}
\end{table}


\paragraph{KDE-Based Manifold Filtering}
We apply a KDE-based filter that assigns generated samples to the data manifold if they exceed a predefined KDE log-likelihood threshold $\tau$. Starting from the data log-likelihood mean, we select $\tau$ via grid search to ensure robust separation between models of differing quality.
Across all experiments, we found $\tau=-300$ to provide a consistent and stable distinction, and we therefore adopt this value throughout\footnote{When the number of generated samples is small, as in our metric stability experiments, it occasionally occurred that no generated sample exceeded the log-likelihood threshold. However, this issue did not arise in our recommended experimental setup using $10{,}000$ samples.}.

\paragraph{Algorithmic Specifics}
The DMMD implementation computes MMD between two sets of embeddings in a memory-efficient block-wise manner, using a block size of $1024$. This value was selected empirically as a good trade-off between memory usage and computational speed. Instead of materializing full kernel matrices, DMMD processes embeddings in chunks and computes the corresponding kernel sums separately before accumulating them into the final estimate. The core routine is compiled with \texttt{@jax.jit}, enabling XLA optimization and parallel execution on GPUs and TPUs. Together, block-wise computation and JIT compilation make DMMD efficient and scalable to large evaluation datasets.
For the experiments with image transformations we used popular torchvision library \citep{torchvision2016}. 


\paragraph{Human Error Rate and Memorization Ratio}


The Human Error Rate and Memorization Ratio scores used in our real-world evaluation of DGMs, following \cite{stein2024exposing}, are reported in \cref{tab:human_mem}.

\begin{table}[h]
\caption{Human Error Rate and Memorization Ratio of various DGMs on CIFAR-10 and ImageNet. Implementations follow \citep{jiralerspong2024feature}. Due to the substantial memory requirements of ImageNet and prior findings indicating that memorization is not a primary concern for this dataset~\citep{stein2024exposing}, the Memorization Ratio is not computed for ImageNet.
}
\label{tab:human_mem}
\resizebox{1\columnwidth}{!}{\begin{tabular}{c@{\,\;}c@{\,\,}c@{\,\,}c@{\;\;\;\;}c@{\,\,}c@{\,\,}c@{\,\,}c}
\toprule
& Model & Hum. Err. Rate$\uparrow$ & Mem. Ratio$\downarrow$ & & Model & Hum. Err. Rate$\uparrow$  \\
\midrule
\multirow{13}{*}{\rotatebox{90}{CIFAR-10}} 
& LSGM-ODE  & 0.437 $\pm$ 0.010 & 0.7293 & \multirow{11}{*}{\rotatebox{90}{\makecell{ImageNet 256$\times$256}}} & DiT-XL-2-guided & 0.330 $\pm$ 0.013  \\
& PFGM++  & 0.436 $\pm$ 0.011 & 0.9730 & & LDM  & 0.309 $\pm$ 0.017  \\
& iDDPM-DDIM  & 0.400 $\pm$ 0.013 & 1.5160 & & DiT-XL-2  & 0.286 $\pm$ 0.016  \\
& StyleGAN-XL & 0.399 $\pm$ 0.012 & 0.7439 & & ADMG-ADMU & 0.269 $\pm$ 0.016  \\
& StyleGAN2-ada & 0.393 $\pm$ 0.012 & 1.0339 & & ADM   & 0.266 $\pm$ 0.016  \\
& BigGAN-Deep  & 0.387 $\pm$ 0.014 & 0.7970 & & ADMG & 0.248 $\pm$ 0.012  \\
& ReACGAN & 0.336 $\pm$ 0.016 & 1.3520 & & RQ-Transformer  & 0.223 $\pm$ 0.012 \\
& MHGAN  & 0.336 $\pm$ 0.015 & 0.6580 & & BigGAN   & 0.210 $\pm$ 0.010 \\
& LOGAN  & 0.206 $\pm$ 0.020 & 0.6760 & & Mask-GIT  & 0.183 $\pm$ 0.016  \\
& WGAN-GP & 0.170 $\pm$ 0.013 & 0.3970 & & GigaGAN  & 0.160 $\pm$ 0.010  \\
& ACGAN-Mod  & 0.148 $\pm$ 0.013 & 0.8610 & & StyleGAN-XL  & 0.153 $\pm$ 0.013  \\
& NVAE  & 0.131 $\pm$ 0.018 & 0.1520 & & & \\
& RESFLOW  & 0.088 $\pm$ 0.008 & 0.4100 & & &  \\
\bottomrule
\end{tabular}}
\end{table} 

\paragraph{Datasets}
CIFAR-10  \cite{krizhevsky2009learning} is a widely used benchmark dataset consisting of $60{,}000$ color images of size $32\times32$, evenly distributed across 10 object classes, with a predefined split of $50{,}000$ training and $10{,}000$ test samples.  

ImageNet (256$\times$256) \cite{deng2009imagenet} is a large-scale visual recognition dataset containing over 1.2 million training images spanning $1{,}000$ object categories, with $50{,}000$ validation images and $100{,}000$ test images. Following common practice in generative modeling, images are center-cropped and resized to $256\times256$. 

 MS-COCO (Microsoft Common Objects in Context) \cite{lin2014microsoft} is a large-scale dataset for object detection, instance segmentation, and image captioning, featuring over $330{,}000$ images (more than $200{,}000$ labeled) with 80 object categories, 1.5 milion object instances, and 5 captions per image.
 
\paragraph{Feature Embeddings}
We compute our metrics in the semantic feature space of DINOv3~\cite{simeoni2025dinov3}, a recent self-supervised Vision Transformer (ViT-L/16). We use DINOv3 as the default backbone to evaluate PALATE with a contemporary high-capacity visual representation, while prior DINOv2-based evaluation work motivates this family of embeddings through their semantic alignment with human perception and cross-dataset robustness. DINOv3 further extends this line with larger-scale training, a high-capacity teacher, Gram-anchoring to prevent feature collapse, and improved dense visual features. To ensure that our conclusions are not driven only by this backbone choice, we repeat representative experiments with DINOv2 embeddings. Figures~\ref{fig:inter_dinov2} and~\ref{fig:stab_dinov2} show the same qualitative trends, suggesting that the reported behavior is primarily induced by the PALATE formulation rather than by a DINOv3-specific effect.

\paragraph{Stable Diffusion Fine-Tuning Details}\label{sec:lora}
We trained a LoRA adapter on top of Stable Diffusion v1.5 text-to-image diffusion model (\texttt{runwayml/}\allowbreak\texttt{stable-}\allowbreak\texttt{diffusion-}\allowbreak\texttt{v1-5}) using the Hugging Face Diffusers script \texttt{train\_}\allowbreak\texttt{text\_}\allowbreak\texttt{to\_}\allowbreak\texttt{image\_}\allowbreak\texttt{lora.py}. Training was performed on the COCO5K subset (image--caption pairs) at a resolution of $512 \times 512$, with batch size 1 and gradient accumulation of 4. We optimized with AdamW using a learning rate of $1 \times 10^{-4}$ for $15{,}000$ update steps. To reduce memory usage, we enabled mixed-precision FP16 training and gradient checkpointing, and saved checkpoints every $1{,}500$ steps. All experiments were executed on a SLURM-managed GPU node with a fixed random seed (42) for reproducibility.



\bibliographystyle{elsarticle-num}
\bibliography{ref}

\end{document}